\documentclass[sigconf,balance=false]{acmart}
\usepackage{graphicx}
\usepackage{tikz}
\tikzset{every picture/.style={/utils/exec={\sffamily}}}
\usepackage{pgfplots}
\pgfplotsset{compat=1.17}
\usepackage{subcaption}
\usepackage[most]{tcolorbox}
\usepackage{algorithm}
\usepackage{algorithmic}
\usepackage{multirow}

\usepackage{popets}
\setcopyright{popets}
\copyrightyear{YYYY}
\acmYear{YYYY}
\acmVolume{YYYY}
\acmNumber{X}
\acmDOI{XXXXXXX.XXXXXXX}
\acmISBN{}
\acmConference{Proceedings on Privacy Enhancing Technologies}
\newtheorem{theorem}{Theorem}
\newtheorem{definition}{Definition}
\newtheorem{remark}{Remark}
\newtheorem{corollary}{Corollary}

\newcommand{\method}{\texttt{Eclipse}}

\newcommand{\red}[1]{\textcolor{red}{#1}}

\newcommand{\blue}[1]{\textcolor{blue}{#1}}

\newcommand{\Alr}{A_{\text{lr}}}
\newcommand{\Aplr}{A_{\text{plr}}}

\newcommand\blfootnote[1]{%
  \begingroup
  \renewcommand\thefootnote{}\footnote{#1}%
  \addtocounter{footnote}{-1}%
  \endgroup
}
\usepackage{environ}
\pgfplotsset{compat = newest}
\usetikzlibrary{arrows,positioning} 
\usetikzlibrary{decorations.text}
\usetikzlibrary{decorations.pathreplacing}
\usetikzlibrary{arrows.meta}
\usetikzlibrary{backgrounds}
\tikzset{
    >=stealth',
    port/.style = {circle, draw, align=center, minimum height=1mm},
    op/.style={
           rectangle,
           rounded corners,
           draw=black, thick,
           text width=3.5em,
           minimum height=1em,
           text centered},
    data/.style={
           rectangle,
           draw=black, thick,
           minimum height=1em,
           text centered},
    area/.style={
           rectangle,
           draw=black, thick,
           minimum height=1em},
    connect/.style={
           ->,
           thick,
           shorten <=2pt,
           shorten >=2pt,}
}

\begin{document}

\title[Edge Private Graph Neural Networks]{Edge Private Graph Neural Networks with Singular Value Perturbation}


\author{Tingting Tang*}
\affiliation{%
  \institution{University of Southern California}
  \city{Los Angeles}
  \state{California}
  \country{USA}}
\email{tangting@usc.edu}

\author{Yue Niu*}
\affiliation{%
  \institution{University of Southern California}
  \city{Los Angeles}
  \state{California}
  \country{USA}
}
\email{yueniu@usc.edu}

\author{Salman Avestimehr}
\affiliation{%
  \institution{University of Southern California}
  \city{Los Angeles}
  \state{California}
  \country{USA}
}
\email{avestime@usc.edu}

\author{Murali Annavaram}
\affiliation{%
 \institution{University of Southern California}
 \city{Los Angeles}
 \state{California}
 \country{USA}}
\email{annavara@usc.edu}


\renewcommand{\shortauthors}{Tang et al.}

\begin{abstract}
Graph neural networks (GNNs) play a key role in learning representations from graph-structured data and are demonstrated to be useful in many applications. However, the GNN training pipeline has been shown to be vulnerable to node feature leakage and edge extraction attacks. This paper investigates a scenario where an attacker aims to recover private edge information from a trained GNN model. Previous studies have employed differential privacy (DP) to add noise directly to the adjacency matrix or a compact graph representation. The added perturbations cause the graph structure to be substantially morphed, reducing the model utility. We propose a new privacy-preserving GNN training algorithm, \method{}, that maintains good model utility while providing strong privacy protection on edges. \method{} is based on two key observations. First, adjacency matrices in graph structures exhibit low-rank behavior.  Thus, \method{} trains GNNs with a low-rank format of the graph via singular values decomposition (SVD), rather than the original graph. Using the low-rank format, \method{} preserves the primary graph topology and removes the remaining residual edges. \method{} adds noise to the low-rank singular values instead of the entire graph, thereby preserving the graph privacy while still maintaining enough of the graph structure to maintain model utility. We theoretically show \method{} provide formal DP guarantee on edges. Experiments on benchmark graph datasets show that \method{} achieves significantly better privacy-utility tradeoff compared to existing privacy-preserving GNN training methods. In particular, under strong privacy constraints ($\epsilon<4$), \method{} shows significant gains in the model utility by up to $46\%$. We further demonstrate that \method{} also has better resilience against common edge attacks (e.g., LPA), lowering the attack AUC by up to $5\%$ compared to other state-of-the-art baselines. 
\end{abstract}

\keywords{Graph Neural Networks, Differential Privacy, Singular Value Decomposition, Node Classification, Edge Privacy Attack \blfootnote{* These authors contributed equally to this work}}
\maketitle

\section{Introduction}
Graph-structured data are ubiquitous in the real world, such as social networks \cite{10.1145/3308558.3313488}, molecules \cite{pmlr-v70-gilmer17a}, and transportation networks \cite{Diao_Wang_Zhang_Liu_Xie_He_2019}. Graph Neural Networks (GNNs) are a family of powerful neural networks that learn representations from such relational data and have achieved state-of-art performance in learning tasks, mainly node classification \cite{DBLP:conf/iclr/KipfW17}, link prediction \cite{10.5555/3327345.3327423} and graph classification \cite{DBLP:conf/iclr/XuHLJ19}. Due to the good performance, GNNs have been deployed in various applications, such as recommender systems \cite{10.1145/3219819.3219890}, drug discovery \cite{rong2020self}, and traffic flow prediction \cite{10.1145/3366423.3380186}. 

Meanwhile, several works have found that GNNs are vulnerable to privacy attacks \cite{he2021stealing, 9833806}, where edges or node features used in training a GNN can be revealed via interaction with the trained GNN during inference. An attacker can send a series of seemingly innocuous inference requests to a trained model and use the model outputs to infer the structure of the graph used to train the GNN~\cite{9833806}, or to extract the node features in the graph~\cite{GNNnodefeatattack}. 
While both the node features and graph structure (often represented by an adjacency matrix) can be sensitive information, in this work we focus on the setting where the node features are public but the graph structure needs to be private. For instance, in a social network a node representation such as the person's education level and city may be public, while that person's social network is private.  In general, graph edges usually encode sensitive information, such as private social relationships among users of a social network, molecular structure of a drug under trade secret, or private user-item interaction history of a user in a recommender system~\cite{privateRS}. Therefore, designing privacy-preserving GNNs to protect private, sensitive edges against such attacks is critical. 

Differential privacy (DP) is a well-known solution to protect against privacy attacks. This technique applies noise to the data while still enabling usage of the perturbed data. In particular, DP has been widely used in machine learning models to protect data and the model. 
One pioneering work, DP-SGD \cite{DPSGD}, applies DP-noise to the gradients during the training process. 
However, the standard DP-SGD algorithm requires gradients of different inputs to be independent. For GNNs with graph data, gradients of each node are affected by its neighboring nodes through node aggregation (See Sec \ref{prelim:gnn}),  making the standard DP-SGD not applicable to GNN settings \cite{GCNRW, GCNNodeDP}. 
Recent works have tailored DP to GNN settings \cite{daigavane2021node, 9833806, 10.1145/3548606.3560705, 287254}. Specifically, to protect graph edges from being leaked via trained models, several edge-level DP algorithms have been proposed \cite{9833806, 10.1145/3548606.3560705, 287254}.
For instance, DPGCN \cite{9833806} directly adds noise to the adjacency matrix and hence the true edge connectivity information is protected. By adding DP noise to the whole adjacency matrix and reconstructing the graph, DPGCN changes the graph structure and reduces the correlation between the true edges in the graph and the graph used during the training process.  DPGCN however has lower privacy-utility tradeoff. The reason for the degraded performance is that binary values in the adjacency matrix (1: edge, 0: non-edge) are very sensitive to noise. The perturbed graph structure can be drastically different from the original graph, even for a moderate privacy budget (See more analysis on Sec \ref{sec:prem:dpgcn}). 
Rather than applying noise to the whole adjacency matrix, LPGNet \cite{10.1145/3548606.3560705} uses a compact representation of the graph structure called cluster degree vector and applies DP-noise only to this structure. 
It further tailors the model architecture to exploit the compact graph information. The graph compaction algorithm along with the accompanied modifications to the model architecture enable LPGNet to be more resilient to noise when applying DP, leading to a better model performance with the same privacy constraints. However, the cluster degree vector preserves only coarse-grain information about the graph, namely the number of neighbors in each cluster for a given node,  and fully discards information of graph connectivity. Thus the model utility is significantly lower compared to GNN trained with a full adjacency matrix.  Furthermore, LPGNet requires a change to the model architecture. In summary, current edge-level DP methods reduce the model utility (F1 score) significantly in favor of privacy and/or require GNN model architecture modifications. 

To address the above limitations, we propose \method{}, a privacy-preserving training methodology for training graph neural networks that utilizes the graph structure to improve the model utility. However, our approach applies DP-noise to the low-rank format of a graph, rather than the adjacency matrix of the graph. In particular, the DP-noise is added to the singular values of the matrix obtained via singular value decomposition (SVD). Hence, the dimension to be considered when applying DP is reduced from $n^2$ to $n$ ($n:$ number of nodes in the graph). Furthermore, as has been observed in prior work~\cite{pmlr-v31-zhou13a}, adjacency matrices in real-world graph data exhibit low-rank structures. That is, most information in the graph can be compressed into a low-dimensional representation. Hence, the DP-noise addition can be limited to just a few ranks.  This key insight allows \method{} to further reduce the dimension that needs to be perturbed by focusing only on a low-rank version of the graph when training GNNs. Unlike LPGNet, \method{} attains the compact format of the graph while still preserving most information, leading to a better privacy-utility tradeoff.

In terms of privacy \method{} protects privacy in two ways. First, with a low-rank graph, \method{} only uses a subset of edges in training, 
and decorrelates the GNN with the rest of the edges.  Then, \method{} further perturbs low-rank singular values and feeds the GNN with a perturbed low-rank reconstruction of the graph during model training, leading to DP on edges in the low-rank graph.
Furthermore, \method{} acts as a plug-in module compatible with existing GNN architectures and only needs to replace the original adjacency matrix input with a perturbed low-rank format. Hence, there is no need to change the model architecture.

We empirically evaluate \method{} on 6 datasets under transductive settings (i.e., different nodes from the same graph are used for training and testing), and 1 dataset under inductive settings(i.e., the nodes in the testing graph are unseen during training). 
In transductive settings, \method{} enjoys much-improved privacy-utility tradeoff over the state-of-the-art baselines, including DPGCN and LPGNet. In particular, under strong privacy constraints ($\epsilon < 4$), \method{} achieves up to $46\%$ higher model utility (F1 score) compared to DPGCN and LPGNet. Even under extreme privacy constraints ($\epsilon < 1$), \method{} still attains up to $5\%$ higher model utility compared to training with node features only (MLP). 

We further evaluate \method{}'s resilience against two common edge attacks: LPA~\cite{he2021stealing} and LINKTELLER~\cite{9833806}. Under LPA attacks, \method{} shows better attack resilience compared to DPGCN and LPGNet. Specifically, with comparable model utility among \method{} and other baselines, the trained model using \method{} results in lower Area Under Receiver Operating Curve (AUC), a well known metric for measuring attack performance (lower AUC means better attack resilience). For instance, on the Cora dataset, with a model utility of around $65\%$, \method{} lowers the attack AUC by $5\%$ compared to LPGNet and DPGCN. Further, attack AUC on \method{} is
at most $5\%$ higher compared to MLP, while the attack AUC on LPGNet and DPGCN can go up to $12.5\%$ and $16\%$ higher than MLP, respectively.
We also provide insights on privacy protection of \method{} on different parts of the graph, including protection on low-degree and high-degree nodes.
\section{Related Work}
In this section, we discuss related works on privacy attacks on GNNs, privacy protection using DP, and graph decomposition.

\textbf{Privacy Attacks on GNNs. } Recent studies have demonstrated that GNNs are vulnerable to privacy attacks on different components of a graph, such as nodes, edges, and graph properties (e.g., graph density) \cite{he2021node,9833806,zhang2022inference}. 
In particular, attacks that recover edges in a graph have been demonstrated. \citet{ijcai2021p516} presents the first model inversion attack on GNNs that reconstructs private edges via white-box access to node attributes, edge density, and the target model parameters. 
\citet{he2021stealing} show black-box link-stealing attacks that can infer the existence of a link between any pair of nodes in the graph with some prior knowledge, such as the target node attributes, extra datasets, etc.
LINKTELLER \cite{9833806} is another important work that proposes a black-box link-stealing attack in a more practical setting based on node influence analysis. And the attack does not require node features when performing attacks.  
More recently, \citet{e98bde528de945d587613f6c34aa1b21} analyze the information leakage based on empirical observation with graph structure reconstruction attacks. They show that additional knowledge of post-hoc feature explanations substantially increases the success rate of these attacks. These works expose the privacy risks of GNNs that use sensitive graph edges during training, and call for effective defense mechanisms to protect the private edges.

\textbf{Differentially Private GNNs. } Differential Privacy (DP) aims to provide formal privacy guarantees for general dataset analysis tasks, and has been widely used in machine learning tasks. It uses a randomized mechanism to perturb the target object and generate indistinguishable outputs for neighboring data.
For neural networks, DP-SGD \cite{DPSGD} acts as a general privacy-preserving training framework that protects models against model inversion or membership inference attacks. It injects noise into gradients during training, and therefore reduces correlations between the final model and the input data. 
However, the standard DP-SGD algorithm requires gradients of different inputs to be independent. For GCNs with graph data, gradients of each node are affected by its neighboring nodes through node aggregation (See Sec \ref{prelim:gnn}), making the standard DP-SGD not applicable to GCN settings \cite{GCNRW, GCNNodeDP}.
In the GNN domain, DP is instead tailored to two scenarios: edge-level DP, which protects the existence of an edge \cite{10.1145/2611523},  and node-level DP, which protects the presence of a node and all the edges connected to it \cite{10.1007/978-3-642-36594-2_26}. 

In the edge-DP setting, \citet{9833806} propose DPGCN, that directly applies DP on the input graph and uses the perturbed graph to train GNNs. As noise is added to the raw graph data, model performance is affected, particularly under stronger privacy constraints.
In contrast to directly perturbing graphs, \citet{10.1145/3548606.3560705} propose LPGNet that adapts a GNN architecture to a low-dimensional graph representation, called \emph{cluster degree vectors}. These vectors are then perturbed via the Laplace mechanism to ensure edge-level privacy. Owing to the reduced dimension when performing DP, LPGNet enjoys better privacy-utility compared to DPGCN. However, model accuracy using LPGNet is capped compared to standard GNNs since limited information is encoded in the degree vectors.
Compared to these approaches, our proposed technique attains a compact format of the graph structure and preserves most information about the graph structure at the same time.

In the node-DP setting, \citet{daigavane2021node} trains GNNs with node-level DP guarantee by extending the standard DP-SGD algorithm and applying privacy amplification by subsampling on bounded-degree graph data. However, this approach requires large privacy budgets to achieve reasonable model accuracy. 
\citet{287254} consider both node-level and edge-level DP, and propose a new GNN architecture GAP to protect both node features and edges. In GAP, node embedding aggregation is decoupled from the standard GNN architecture. They apply noise to the output of the node embedding aggregation (with the adjacency matrix), therefore achieving both edge-level and node-level DP. 

There are alternative approaches to protect data privacy during training and inference, such as Secure Multi-Party Computing (MPC) \cite{wangCharacIspass, wang2022mpcpipe, wang2023compacttag}, Coded Computing \cite{yu2019lagrange, tang2022avcc}, and Trusted Execution Environment (TEE) \cite{slalom19,narra21tee,hashemi21tee}. However, these approaches target different settings from ours, and cannot protect against edge attacks exploiting correlations among model outputs. Thus, in this work, we do not consider these approaches and instead focus on the DP-based approach to provide edge privacy.

\textbf{Applying SVD on GNNs. } Singular value decomposition (SVD) is a commonly used technique in data processing. 
By encoding the most important information into a low-dimensional space, it shows significant benefits in reducing complexity in DNNs \cite{DBLP:journals/popets/NiuAA22, prism_tmlr, delta_arxiv, Lowrank_BMVC_2014, Lowrank_ICLR_2016}. Recent endeavors have also shown its potential in graph tasks, especially for defending against adversarial attacks on graph data. 
\citet{10.1145/3336191.3371789} demonstrated that small perturbation by an adversarial attack is only reflected on high-rank residual components, whereas the low-rank approximation is minimally affected. They exploit this observation to enable GNN training with a low-rank graph, which is more immune to adversarial attacks. 
\citet{10.1145/3459637.3482299} aim to speed up robust structure learning, which is another method to improve the robustness of GNNs against adversarial attacks. Their approach exploits the low-rank property of the adjacency matrix of a graph as prior knowledge, and decouples the adjacency matrix into a low-rank component and a sparse one, and learns by minimizing matrix rank. Despite the promising applications of SVD demonstrated in improving the robustness of GNNs, it is rarely seen employing SVD in the privacy domain when training GNNs. In this work, we aim to use SVD to study the graph structure and exploit its potential in privacy-preserving GCN training.

\section{Preliminaries}

\subsection{Graph Neural Networks}\label{prelim:gnn}

GNNs learn a node embedding for each node by distilling the high-dimensional information about a node’s graph neighborhood into a low-dimensional embedding. The learned node embeddings can then aid downstream graph learning tasks, such as node classification, link prediction, and clustering. In this work, we focus on graph convolutional neural network (GCN), a prominent GNN type widely used for semi-supervised graph classification tasks on graphs. Consider an unweighted graph $\mathcal{G} = (\mathcal{V}, \mathcal{E})$, where $\mathcal{V}$ denotes the set of nodes ($n = |\mathcal{V}|$), and $\mathcal{E}$ the set of edges which can be represented by an adjacency matrix $A \in \{0,1\}^{n \times n}$, with $A_{ij} = 1$ indicating the presence of an edge between node $i$ and node $j$. GCN takes as input the adjacency matrix $A$ and the node feature matrix $X \in \mathbb{R}^{N \times d}$, where each row of $X$ corresponds to a $d$-dimensional feature vector of a node in $\mathcal{V}$. A $k$-layer GCN consists of a sequence of $k$ graph convolutional layers. To obtain a new embedding for each node in the graph, each hidden layer aggregates the embeddings of the node’s neighbors output by the previous layer. The $l$-th layer can be formally described as: 
\begin{equation}\label{eq:HiddenLayer}
    H^{l+1} = \sigma(\tilde{A}H^{l}W^{l}),
\end{equation}
where $\tilde{A}$ is normalized adjacency matrix, $\sigma$ is the nonlinear activation function such as ReLU, $H^{l}$ is the node embeddings in the $l$-th layer where $H^{0}=X$ is the initial node feature, and $W^{l}$ is the learnable weight parameters. 
$\tilde{A}H^l$ denotes the feature aggregation step. The normalized $\tilde{A}$ can be calculated as

\begin{equation}\label{eq:Anormalization1}
    \text{FirstOrderGCN:}  \quad \tilde{A} = I + D^{-0.5}\cdot A \cdot D^{-0.5},
\end{equation}
or 
\begin{equation}\label{eq:Anormalization2}
    \text{AugNormAdj:} \quad \tilde{A} = (D+I)^{-0.5}\cdot A \cdot (D+I)^{-0.5},
\end{equation}
where $D = \texttt{Diag}(d_0, d_1, \cdots, d_{|V|-1}), d_i$ denotes degree of $i$-th node. Depending on datasets and models, different methods may prefer different normalization methods. 

\begin{figure*}[!htb]
    \centering
    \begin{tikzpicture}
        \node[](GRAPH){\includegraphics[width=.1\textwidth]{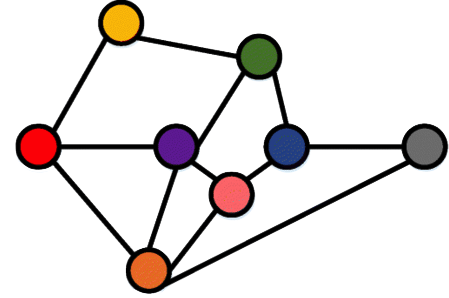}};
        \node[data, draw=black!30, very thin, fill=black!10, text width=10mm, right=5mm of GRAPH, yshift=5mm](A){\scriptsize adj. matrix \\ $A$};
        \node[data, draw=black!30, very thin, fill=black!10, text width=10mm, right=5mm of GRAPH, yshift=-5mm](X){\scriptsize feature \\ $X$};
        \node[op, draw=red!30, fill=red!30, text width=12mm, minimum height=5mm, right=10mm of X ](GNN1){\scriptsize GCN layer};
        \draw[decorate, decoration={brace, raise=-4pt, amplitude=4pt}, draw=red!70] (GNN1.south west) -- +(15mm, 0mm);
        \node[below=3mm of GNN1](){\scriptsize GCN: $H^{l+1} = \sigma (\tilde{A} \cdot H^l \cdot W)$};
        \node[op, draw=red!30, fill=red!30, text width=12mm, minimum height=5mm, right=10mm of GNN1](GNN2){\scriptsize GCN layer};
        \node[data, draw=black!30, very thin, fill=black!10, text width=5mm, right=5mm of GNN2](OUT){out};

        \draw[->] (GRAPH.east) to [out=0, in=180] (A.west);
        \draw[->] (GRAPH.east) to [out=0, in=180] (X.west);
        \draw[->] (X.east) to [out=0, in=180] node[above] {\tiny $H^l$} (GNN1.west);
        \draw[->] (GNN1.east) to [out=0, in=180] node[above] {\tiny $H^{l+1}$} (GNN2.west);
        \draw[->] (GNN2.east) to [out=0, in=180] (OUT.west);

        \draw[->] (A.east) -- +(15mm, 0mm) to [out=0, in=90] (GNN1.north);
        \draw[->] (A.east) -- +(39mm, 0mm) to [out=0, in=90] (GNN2.north);
    \end{tikzpicture}
    \vspace{-2mm}
    \caption{A typical 2-layer GCN. Each GCN layer takes as input node embedding, $H^l$, and performs aggregation $\tilde{A}\cdot H^l$ with normalized adjacency matrix $\tilde{A}$. Then, a weight matrix $W$ is applied to the aggregated embeddings.}
    \label{fig:2layergcn}
\end{figure*}
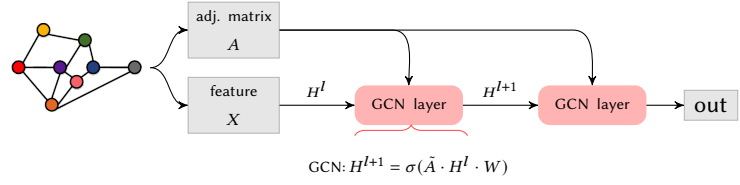

Node embeddings from the last layer are transformed into probability scores via a softmax layer for the node classification task. A typical 2-layer GCN is illustrated in Figure \ref{fig:2layergcn}.

Node-level classification can occur in two settings: \emph{transductive and inductive} settings. In the \emph{transductive setting}, the training and testing are performed on the same graph but with different nodes. While in the \emph{inductive setting}, the graph in testing is distinct from the one in training.

\subsection{Differential Privacy}
Differential privacy (DP) has been known as a rigorous privacy-preserving methodology and has been widely adopted in privacy-sensitive applications \cite{10.1007/11681878_14}.

\begin{definition}\label{def:dp}
$(\epsilon, \delta)$-DP: A randomized mechanism $\mathcal{M}: \mathbb{D}\rightarrow \mathbb{R}$ satisfies $(\epsilon, \delta)$-DP if for any two adjacent inputs $x, x'\in \mathbb{D}$ and any output $\mathbb{S} \subseteq \mathbb{R}$, it holds that 

\begin{center}
    $P[ \mathcal{M}(x) \in \mathbb{S} ] \leq e^{\epsilon} \cdot P[ \mathcal{M}(x') \in \mathbb{S} ] + \delta $.
\end{center}
    
\end{definition}

Depending on the applications, the concept of adjacent data varies. 
In our case, we target adjacency matrix protection of undirected unweighted graphs in graph neural networks. We define adjacent matrices $A, A'$ if \emph{they only differ in one edge}. That is, the graph represented by $A'$ is obtained by adding/removing one edge in the graph represented by $A$. Specifically, if two graphs differ on one edge between node $i$ and node $j$, for the \emph{symmetric} adjacency matrices $A$ and $A'$, they differ at $(i,j)$ and $(j,i)$ (i.e., $A(i,j)=A(j,i)=1$, $A'(i,j)=A'(j,i)=0$).
By replacing $x$ with $A$ in Def \ref{def:dp}, we define edge-level DP as follows.

\begin{definition}\label{def:edp}
$(\epsilon, \delta)$-edge DP: A randomized mechanism $\mathcal{M}$ satisfies $(\epsilon, \delta)$-edge DP if for any two adjacent graphs $A, A'$ and any output $\mathbb{S} \subseteq \mathbb{R}$(range of $\mathcal{M}$), it holds that 

\begin{center}
    $P[ \mathcal{M}(A) \in \mathbb{S} ] \leq e^{\epsilon} \cdot P[ \mathcal{M}(A') \in \mathbb{S} ] + \delta $.
\end{center}
    
\end{definition}

However, DP often sacrifices performance for strong privacy constraints (small $\epsilon$). Balancing model performance and privacy of the adjacency matrix is a challenge when working with GNNs.

\subsection{Edge-Level DP on GNN}\label{sec:prem:dpgcn}
In this section, we describe two typical approaches to edge-level DP on GNN: DPGCN and LPGNet. 

DPGCN \cite{9833806} directly adds noise to the adjacency matrix $A$ and reconstructs the graph as 
\begin{equation*}
    \hat{A} = \texttt{binary}(A + \texttt{Laplacian}(0, \sigma)),
\end{equation*}
where $\sigma$ controls the variance of a Laplacian noise matrix that has the same dimension as $A$. The noise added is determined by the $\epsilon$ value selected. The smaller the $\epsilon$ larger is the noise.  The \texttt{binary} op simply sorts values in $A$ and sets the first $|\mathcal{E}|$ (number of edges in A) to 1 and zeros to other values. Therefore, the perturbed graph $\hat{A}$ still has similar sparsity as the original $A$, but with different connections.
Since $A$ consists only of binary values, the resulting $\hat{A}$ can be drastically different with a certain level of noise added. That is the main reason that DPGCN's performance quickly drops when the privacy budget decreases.

LPGNet mitigates the issue by extracting a much-compressed representation of the graph called cluster degree vectors, $D$. Specifically, for each node, LPGNet computes the number of its neighbors on each cluster (i.e., nodes with the same label), and perturbs the degree vectors as
\begin{equation*}
    \hat{D} = D + \texttt{Laplacian}(0, \sigma),
\end{equation*}
where each row in $D\in\mathbb{Z}^{n\times L}$ denotes the number of the node's neighbors in each cluster, $L$ indicates the number of clusters/labels, $\sigma$ controls the variance of the Laplacian noise matrix that has the same dimension as $D$. By using the compressed representation, LPGNet discards detailed connection information and sacrifices model performance with an improved privacy guarantee. 
\section{Problem Settings and Threat Model}

\textbf{Problem Settings. }
We target a node classification problem over a graph when node features are public and edges are private. The problem arises in real scenarios where edges may represent sensitive social or financial relationships. Such relational data can significantly enhance GNNs to learn representations and relations from public node features \cite{10.1145/3548606.3560705, DPGCN}. For instance, social networks such as LinkedIn and Twitter allow users to hide their connections for strong privacy protection (edges), while still maintaining a public profile (node features). In these scenarios, private connections may contain sensitive information not present in node features.

The entity with node features trains GNNs with additional information from the graph provider. After training, the model is deployed, and receives queries for classifying certain nodes (e.g., identifying user groups). 
However, even if the GNN model is trained using a private adjacency matrix, it is possible that the adjacency information can be leaked by interacting with the model. For instance, well-known attack methods such as LPA \cite{he2021stealing} and LINKTELLER~\cite{9833806} can effectively estimate the graph edges by sending a series of inference queries and analyzing the model's outputs. 

Therefore, this work aims to \emph{train a GCN model with good utility using private graph data, while still maintaining strong protection on the graph during/after training.} 

\textbf{Threat Model.  }
We assume an adversary has access to node features and models' API. It learns certain prior knowledge about a target graph through node features. It can also query models with node features with an unlimited number of queries.  
However, the adversary has no access to the private graph structure $A$. 

\section{Proposed Method}\label{sec:method}
Based on the data shown in prior works~\cite{9833806}, it is best to avoid applying DP noise to the adjacency matrix $A$, as the noise alters the graph structure significantly leading to performance drop. On the other hand ignoring the graph structure entirely is also unacceptable. In fact, prior works have demonstrated that using a distilled version of a graph \cite{10.1145/3548606.3560705} improves model performance. However, the performance gap is still high compared to a non-private GCN training approach. The reason is that very limited information is preserved when converting the graph into a compact format as in \cite{10.1145/3548606.3560705}. Based on these observations this work presents a new approach called \method{} that trains a GCN with a low-rank format of a graph. The low-rank format preserves as much information about a graph as possible but with a low-dimensional format. 

We first demonstrate that the difference between $A$ and $A'$, measured as $l_2$ norm, can be transferred to their singular values, therefore reducing the dimension of DP-noise from $n^2$ to $n$. Thus the amount of noise added for a given $\epsilon$ privacy budget can be reduced.  Furthermore, we observe that high-rank singular values differ more compared to the low-rank part.  With the observation, \method{} trains GCNs with a low-rank approximation of $A$. To achieve DP, \method{} only needs to add noise to the low-rank singular values.  Owing to the reduced dimension, \method{} achieves high accuracy while maintaining strong privacy protection.

\subsection{Graph Decomposition}\label{sec:method:decomp}
In this section, we describe the graph decomposition that we use in our scheme. The essence of the graph decomposition in \method{} is to reduce dimensions for applying DP noise, therefore leading to better privacy-utility trade-offs when applying DP. We then demonstrate how the l2 norm difference between $A$ and $A'$ adjacency matrices is reflected in their singular values.

Given graph $A \in \{0,1\}^{n\times n}$ and $A' \in \{0,1\}^{n\times n}$ (with one edge removed from $A$), we first decompose $A$ as 

\begin{equation}\label{eq:ADecomp}
    A \xrightarrow[]{SVD} U \cdot \texttt{Diag}(s) \cdot V,
\end{equation}
where $S = \texttt{Diag}(s)$ maps singular values $s$ to a 2-D diagonal matrix, $U, V$ denotes the principal components of $A$. In general, for a matrix, SVD finds a set of orthogonal basis vectors, $U, V$, to represent the matrix. The singular values obtained from SVD reflect the amount of information of the matrix contained along corresponding basis vectors. With this notion, we apply SVD to the adjacency matrix $A$, and analyze its bases and each basis vector's importance.

\textbf{Decomposition on $A'$.}
If performing a normal SVD on $A'$ as in Eq(\ref{eq:ADecomp}), it ends with three matrices that may be different from the ones obtained from $A$.  Hence applying DP to protect graph connectivity requires adding noise to all the three corresponding matrices, $U,V$, and $S$, which goes against our objective of applying noise to a reduced dimension graph information. 

\begin{figure}[!htb]
    \centering
    \begin{tikzpicture}
        \begin{axis} [
            ybar=1pt, bar width=2pt, ticklabel style={font=\scriptsize},
            width=.9\linewidth, height=.4\linewidth,
            xmin=0, xmax=21,
            xtick={1, 5, 10, 15, 20}, 
            xticklabels={1, 5, 10, 15, 20},
            xlabel={\scriptsize Index},
            ylabel={\scriptsize Similarity}, ymin=0.2,
            ytick={0, 0.5, 1}, yticklabels={0, 0.5, 1},
            ymajorgrids, major grid style={line width=.2pt, draw=black!20, dashed},
            legend columns=1,
            legend style={at={(1.1,1.0)}, anchor=north },
            legend entries={\tiny $U$, \tiny $V$}
        ]
            \addplot[fill=blue!30, draw=blue!60] coordinates {
                (1, 1) (2, 1) (3, 1) (4, 0.999) (5, 0.999)
                (6, 1) (7, 0.999) (8, 1) (9, 1) (10, 1)
                (11, 1) (12, 1) (13, 1) (14, 0.999) (15, 1)
                (16, 1) (17, 1) (18, 1) (19, 1) (20, 1)
            };
				
            \addplot[fill=magenta!30, draw=magenta!60] coordinates {
                (1, 1) (2, 1.001) (3, 1.001) (4, 1) (5, 1)
                (6, 1.001) (7, 1.001) (8, 1.001) (9, 1.001) (10, 1.001)
                (11, 1.001) (12, 1.001) (13, 1.001) (14, 1.001) (15, 1.001)
                (16, 1.001) (17, 1.001) (18, 1.001) (19, 1.001) (20, 1.001)
            };
            
        \end{axis}
    \end{tikzpicture}   
    \vspace{-3mm}
    \caption{Cosine similarity of principal basis vectors in the Chameleon dataset.}
    \label{fig:cos}
\end{figure}
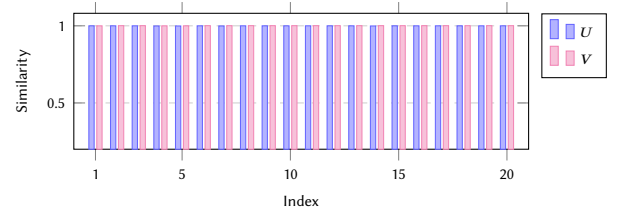

One important observation is that given $A'$ and $A$ only differ at one edge, their principal bases (i.e., $U, V$) after the SVD operation tend to be similar. 
Specifically, we randomly remove one edge in $A$, and compute the cosine similarity of every principal basis between $A$ and $A'$ as
\begin{equation*}
    \texttt{sim}_{U}  = \frac{\left\langle U_i, U'_i  \right\rangle}{\left\| U_i \right\|\cdot \left\| U'_i \right\|}, \quad \texttt{sim}_{V}  = \frac{\left\langle V_i, V'_i  \right\rangle}{\left\| U_i \right\|\cdot \left\| U'_i \right\|},
\end{equation*}
where $U', V'$ are the principal bases from $A'$ via normal SVD, $U_i=U(:,i), V_i=V(:,i)$, so as $U'_i, V'_i$.
Large similarity indicates two vectors are close. For a graph $A$, we denote a $A'$ the worst case if it has the smallest average cosine similarity to $A$ across all principal components. 
For instance, Figure \ref{fig:cos} shows the cosine similarity of the top 20 principal basis vectors on the Chameleon dataset~\cite{sen2008collective}. 
Owing to the closeness of $A$ and $A'$, their principal components have cosine similarity close to $1$. The difference between $A$ and $A'$ is mainly on their singular values. 
More empirical evidence on other datasets is shown in Appendix \ref{app:cosine}. 
This observation motivates us to apply DP only to the singular values, rather than $n^2$ elements in the original graph, and leave the principal basis untouched. Our privacy analysis in Section~\ref{sec:privacy:unreal} discusses the implication of privacy protection beyond this assumption. 

Therefore, we can let $A'$ be \emph{anchored} to the principal basis of A as

\begin{equation}\label{eq:A1Decomp}
    A' \rightarrow U \cdot S' \cdot V
\end{equation}

The essence of Eq(\ref{eq:A1Decomp}) is to show that $A$ and $A'$ can share the same basis, while their difference is reflected only in their singular value matrices via proper decomposition.
Then we can obtain $S'$ as

\begin{equation}\label{eq:S1}
    S' = U^T \cdot A' \cdot V^T.
\end{equation}

\textbf{Low-Rank Structure. }
We further observe that graph matrix A exhibits a low-rank structure, with singular values decaying quickly at the beginning as shown in Figure \ref{fig:observation:singular}. The low-rank structure is rooted in real-world observations. For instance, in social networks, users close to each other tend to have similar connections, leading to highly correlated vectors in the adjacency matrix. 
\pgfplotstableread[col sep=comma]{cora.csv}{\loadedtableA}
\pgfplotstableread[col sep=comma]{chameleon.csv}{\loadedtableB}
\pgfplotstableread[col sep=comma]{citeseer.csv}{\loadedtableC}
\pgfplotstableread[col sep=comma]{facebook.csv}{\loadedtableD}
\pgfplotstableread[col sep=comma]{twitch.csv}{\loadedtableE}

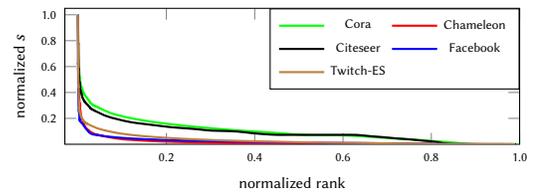
\begin{figure}[!htb]
    \centering
    \begin{tikzpicture}
        \begin{axis}[
            width=.9\linewidth, height=.4\linewidth,
            xlabel={\scriptsize normalized rank},
            xmin=-0.03, xmax=1,
            xtick={0.2, 0.4, 0.6, 0.8, 1.0 },
            xticklabels={0.2, 0.4, 0.6, 0.8, 1.0},
            ymin=0, ymax=1.05,
            ylabel={\scriptsize normalized $s$},
            ytick={0.2, 0.4, 0.6, 0.8, 1.0},
            yticklabels={0.2, 0.4, 0.6, 0.8, 1.0 },
            ticklabel style={font=\tiny},
            legend columns=2,
            legend style={at={(1.0,1.0)}, anchor=north east},
            legend entries={\tiny Cora, \tiny Chameleon, \tiny Citeseer, \tiny Facebook, \tiny Twitch-ES}
        ]

        \addplot[thick, green, smooth] table [x=indx, y=value, col sep=comma] {\loadedtableA};

        \addplot[thick, red, smooth] table [x=indx, y=value, col sep=comma] {\loadedtableB};
        
        \addplot[thick, black, smooth] table [x=indx, y=value, col sep=comma] {\loadedtableC};
        
        \addplot[thick, blue, smooth] table [x=indx, y=value, col sep=comma] {\loadedtableD};
        
        \addplot[thick, brown, smooth] table [x=indx, y=value, col sep=comma] {\loadedtableE};

        \end{axis}
    \end{tikzpicture}
    \vspace{-3mm}
    \caption{Distribution of singular values (normalized) in typical graph data. Singular values decay quickly, indicating that the graph matrix exhibits a highly low-rank structure.}
    \label{fig:observation:singular}
\end{figure}

With such an observation, we can approximate graph $A$ with a low-rank version $A_{\text{lr}}$ as

\begin{equation}\label{eq:lrA}
    A_{\text{lr}} = U(:, 1:r) \cdot \texttt{Diag}(s(1:r)) \cdot V(1:r, :),
\end{equation}
where $U(:, 1:r), V(1:r, :)$ denotes the first $r$ columns and rows in $U, V$.

According to Eq(\ref{eq:A1Decomp}), we can also write the low-rank version of $A'$ as 

\begin{equation}\label{eq:lrA1}
    A'_{\text{lr}} = U(:, 1:r) \cdot \texttt{Diag}(s'(1:r)) \cdot V(1:r, :),
\end{equation}
with the same basis as in $A$, $s'$ denote the diagonal elements in $S'$.

Approximating the graph with a low-rank format is beneficial in two aspects. First, with a low-rank graph, many edges from the original graph are removed. Therefore, training GCNs with the low-rank graph de-correlates the final model from these edges. 
Second, with the low-rank approximation, we further reduce the dimension to be considered and only need to perturb the first $r$ singular values when applying DP.

\subsection{DP on Singular Values}\label{sec:method:dp}
With the notation that $A$ and $A'$ can share principal bases while differing in their singular values, we use a $\ell_2$-sensitivity to define the difference between $A$ and $A'$ as 
\begin{equation}\label{eq:sensitivity}
    \Delta = \max_{A, A'} \left \| s(1:r) -  s'(1:r)\right \|_F
\end{equation}

With the $\ell_2$-sensitivity, we apply a standard Gaussian mechanism to perturb the singular values as follows:

\emph{Gaussian Mechanism on Singular Values. }
For a low-rank approximation $A_{\text{lr}}$ with $r$ bases, a noise vector $g\in \mathbb{R}^r$ following normal distribution is generated as $g \sim \texttt{Normal}(0, \sigma)$. Then the noise vector $g$ is added to the singular values as $\tilde{s} = s + g$.

The noise parameter $\sigma$ depends on privacy constraint $\epsilon$ and sensitivity $\Delta$.
The following lemma shows the sensitivity, $\Delta$, is well-bounded given $A$ and $A'$ share the same principal bases. 
\begin{lemma}\label{lemma:Delta}
Let $s = \texttt{Diag}(U^T \cdot A \cdot V^T)$ and $s' = \texttt{Diag}( U^T \cdot A' \cdot V^T)$, where $U$ and $V$ are principal bases obtained from SVD. Given $A$ and $A'$ share principal bases, we have $\Delta \leq \sqrt{2}$.
\end{lemma}
\begin{proof}
Knowing that $A$ and $A'$ differ in one edge, 
\begin{equation*}
    \left \| A-A' \right \|_2 = \sqrt{2}.
\end{equation*}
The goal is to bound the following quantity:
\begin{equation*}
    \left \| s(1:r) -  s'(1:r)\right \|_F.
\end{equation*}
Given $A$ and $A'$ share principal bases $U$, $V$, and $U$, $V$ are unitary matrices,
\begin{equation*}
\begin{split}
    \left \| s(1:r) -  s'(1:r)\right \|_F &= \sqrt{\sum_{i=1}^r (s(i)-s'(i))^2}\\ &\leq \sqrt{\sum_{i=1}^n (s(i)-s'(i))^2} = \left \| s-s' \right \|_F \\
    &= \left \| U \cdot \texttt{Diag}(s-s') \cdot V \right \|_F = \left \| A-A' \right \|_F \\
    &= \sqrt{2},
\end{split}
\end{equation*}
which concludes the proof.
\end{proof}
\noindent Then we can compute $\sigma$ given a privacy budget $\epsilon$ (See Privacy Analysis in Sec \ref{sec:privacy}).
A high privacy constraint (small $\epsilon$) or higher sensitivity requires large noise (high $\sigma$) for singular values.

\begin{figure*}[!htb]
    \centering
    \begin{tikzpicture}
        \node[](GRAPH){\includegraphics[width=.1\textwidth]{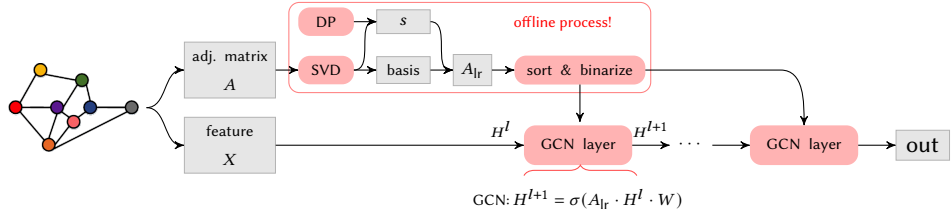}};
        \node[data, draw=black!30, very thin, fill=black!10, text width=10mm, right=5mm of GRAPH, yshift=5mm](A){\scriptsize adj. matrix \\ $A$};
        \node[op, draw=red!30, fill=red!30, text width=5mm, right=3mm of A](SVD){\scriptsize SVD};
        \node[data, draw=black!30, very thin, fill=black!10, text width=5mm, right=3mm of SVD](BASIS){\scriptsize basis};
        \node[data, draw=black!30, very thin, fill=black!10, text width=5mm, above=3mm of BASIS](S){\scriptsize $s$};
        \node[op, draw=red!30, fill=red!30, text width=5mm, left=3mm of S](DP){\scriptsize DP};
        \node[data, draw=black!30, very thin, fill=black!10, text width=3mm, right=3mm of BASIS](A1){\scriptsize $A_{\text{lr}}$};
        \node[op, draw=red!30, fill=red!30, text width=15mm, right=3mm of A1](BIN){\scriptsize sort \& binarize};
        \node[data, draw=black!30, very thin, fill=black!10, text width=10mm, right=5mm of GRAPH, yshift=-5mm](X){\scriptsize feature \\ $X$};
        \node[op, draw=red!30, fill=red!30, text width=12mm, minimum height=5mm, right=33mm of X ](GNN1){\scriptsize GCN layer};
        \draw[decorate, decoration={brace, raise=-4pt, amplitude=4pt}, draw=red!70] (GNN1.south west) -- +(15mm, 0mm);
        \node[below=2mm of GNN1](){\scriptsize GCN: $H^{l+1} = \sigma (A_{\text{lr}} \cdot H^l \cdot W)$};
        \node[right=5mm of GNN1](MORE){\scriptsize $\cdots$};
        \node[op, draw=red!30, fill=red!30, text width=12mm, minimum height=5mm, right=5mm of MORE](GNN2){\scriptsize GCN layer};
        \node[data, draw=black!30, very thin, fill=black!10, text width=5mm, right=5mm of GNN2](OUT){out};

        \node[rounded corners, draw=red!50, text width=46mm, minimum height=12mm, xshift=53mm, yshift=8mm](BOX){};
        \node[above=2mm of A1, xshift=12mm](){\scriptsize \red{offline process!}};

        \draw[->] (GRAPH.east) to [out=0, in=180] (A.west);
        \draw[->] (GRAPH.east) to [out=0, in=180] (X.west);
        \draw[->] (A.east) to [out=0, in=180] (SVD.west);
        \draw[->] (SVD.east) to [out=0, in=180] (BASIS.west);
        \draw[->] (SVD.east) to [out=0, in=180] (S.west);
        \draw[->] (DP.east) to [out=0, in=180] (S.west);
        \draw[->] (S.east) to [out=0, in=180] (A1.west);
        \draw[->] (BASIS.east) to [out=0, in=180] (A1.west);
        \draw[->] (A1.east) to [out=0, in=180] (BIN.west);

        \draw[->] (X.east) to [out=0, in=180] (GNN1.west) node[above, xshift=-3mm] {\tiny $H^l$};
        \draw[->] (GNN1.east) to [out=0, in=180] node[above] {\tiny $H^{l+1}$} (MORE.west);
        \draw[->] (MORE.east) to [out=0, in=180] (GNN2.west);
        \draw[->] (GNN2.east) to [out=0, in=180] (OUT.west);

        \draw[->] (BIN.south) -- +(0mm, -5mm);
        \draw[->] (BIN.east) -- +(18mm, 0mm) to [out=0, in=90] (GNN2.north);
    \end{tikzpicture}
    \vspace{-2mm}
    \caption{GCN with low-rank graph. \method{} first applies SVD to obtain singular values of a graph, and perturbs singular values with Gaussian noise. \method{} then reconstructs a low-rank version of $A$ by only keeping most principal components. To align with original format of the adjacency matrix, \method{} applies a binary quantization before feeding $A_{\text{lr}}$ to a GCN.}
    \vspace{-2mm}
    \label{fig:method}
\end{figure*}

\subsection{GCN Training and Testing}\label{sec:method:training}
\method{} follows the same procedure as standard GCNs, except that a perturbed low-rank adjacency matrix is fed to the model. 
As shown in Figure \ref{fig:method}, during GCN training, we first obtain a perturbed low-rank format, $A_{\text{lr}}$. 
Since the adjacency matrix is used as an input and does not change during training, we only need to perturb it once and fix it throughout the whole training period. 
To avoid additional overhead when performing SVD, we compute $A_{\text{lr}}$ offline so that it does not affect the training time. 

Considering the original adjacency matrix $A$ only consists of binary values, we further align $A_{\text{lr}}$ with the standard format by binary quantization. Specifically, given the number of edges, $|\mathcal{E}|$, in the original graph, we sort elements in $A_{\text{lr}}$ and set the first $|\mathcal{E}|$ largest values to 1, and zero other values in order to preserve the sparsity of the adjacency matrix. The resulting adjacency matrix is denoted as $A_{\text{plr}}$.
In addition, we further consider perturbing the number of total edges $E$. To that end, we distribute the privacy budget into $\epsilon_e$ and $\epsilon_{\text{lr}}$, where $\epsilon_e$ is for the number of edges, and $\epsilon_{\text{lr}}$ is for the low-rank adjacency matrix.
By the composition rule of DP, the total privacy budget on the graph edges is still $\epsilon$.
Algorithm \ref{alg:method} summarizes the perturbation mechanism in our method \method{}.
\begin{algorithm}[!htb]
\caption{Perturbation mechanism in \method{}}
\label{alg:method}
\begin{algorithmic}[1]
\REQUIRE{Graph $\mathcal{G} = (\mathcal{V},\mathcal{E})$ as symmetric adjacency matrix $A$, rank $r$, privacy parameters $\epsilon, \delta$, sensitivity $\Delta = \sqrt{2}$, randomization generator}
\ENSURE{the perturbed low rank outcome $\Aplr$}
\STATE $\epsilon_{e} = 0.01\epsilon$ 
\STATE \blue{// Distribute privacy budget}
\STATE $\epsilon_{\text{lr}} = \epsilon - \epsilon_{e}$
\STATE $E = |\mathcal{E}|$
\STATE $\tilde{E} = E + \mathtt{Laplacian}(0, \frac{1}{\epsilon_{e}})$
\STATE $\tilde{E} = \lfloor \tilde{E} \rfloor$
\STATE $U, \mathtt{Diag}(s), V = SVD(A)$
\FOR {$i =1, \cdots, r$}
    \STATE $\tilde{s}(i) = \tilde{s}(i) + \mathtt{Normal}(0, \frac{\Delta\sqrt{2\ln({1.25/\delta})}}{\epsilon_{\text{lr}}})$
\ENDFOR
\STATE $A_{\text{lr}} = U(: ,1:r) \cdot \mathtt{Diag}(\tilde{s}(1:r)) \cdot V(1:r,:)$
\STATE $A_{\text{tr}} = \mathtt{UppTriMatrix}(\Alr,1)$ \\
\STATE \blue{// Upper triangular matrix on $1^{st}$ superdiagonal}
\STATE \textit{top}\_\textit{indices} = argmax($A_{\text{tr}}, \tilde{E}$)
\STATE \blue{// Get the indices of top $\tilde{E}$ values}
\STATE $A_{\text{tr}}[\textit{top}\_\textit{indices}] = 1$, $A_{\text{tr}}[\neg\textit{top}\_\textit{indices}] = 0$
\STATE $\Aplr = A_{\text{tr}} + A_{\text{tr}}^{T}$
\RETURN $\Aplr$
\end{algorithmic}
\end{algorithm}

\textbf{Transductive Settings. } In a transductive setting, as the same graph is used in training and testing, we only need to apply the SVD and DP offline process once and obtain $A_{\text{lr}}$ for both training and testing. That is, given nodes chosen in training and testing, we extract the edge information from $A_{\text{lr}}$ that covers the chosen nodes. 

\textbf{Inductive Settings. } In an inductive setting, we apply the SVD and DP offline process separately on the training and testing graph and obtain two perturbed low-rank graphs.

\section{Privacy Analysis}\label{sec:privacy}

In this section, we analyze privacy protection using \method{} from two aspects: formal privacy analysis with DP and additional privacy protection via low-rank decomposition. 
We further discuss the effectiveness of \method{} in real-world scenarios. 

\subsection{Formal DP Analysis}

Based on the observations on real-world datasets (e.g. Figure \ref{fig:cos}), it is reasonable to assume that two adjacent graphs share the same principal bases. The formal privacy analysis is built based on this assumption. We also discuss the implication of privacy protection beyond this assumption in Sec \ref{sec:privacy:real}.

Given an adjacency matrix $A$, \method{} decomposes $A$ into principal bases $U, V$, and singular values $s$ (See Eq(\ref{eq:ADecomp})). 
As the principal bases are shared with its neighboring graph $A'$, \method{} achieves $(\epsilon, \delta)-$ edge DP by only perturbing singular values.  

\setcounter{theorem}{0}
\begin{theorem}
\label{theorem:dpsvd}
Given $A$ and $A'$ share the same principal bases, with Gaussian noise,
\begin{center}
    $ \texttt{Normal} \left (0, \Delta\sqrt{2\ln({1.25/\delta})} / \epsilon_{\text{lr}} \right )$,
\end{center} 
added to singular values;
and Laplacian noise, 
\begin{center}
    $\texttt{Laplacian} \left (0,1 / \epsilon_{\text{e}} \right)$,
\end{center}
added to the edge count, the perturbed low-rank $\Aplr$ using \method{} satisfies $(\epsilon,\delta)$-edge DP with $\epsilon = \epsilon_{\text{lr}} + \epsilon_{\text{e}}$.
\end{theorem}
The proof follows the standard Gaussian mechanism, the composition theorem, and the post-processing rule of DP algorithm \cite{DPFoundation}. The detailed proof is given in Appendix \ref{appx:thm1}. 

\begin{corollary}
\label{coro:model}
Given node features and the adjacency matrix with \emph{independent} distributions, the GCN model trained by Algorithm \ref{alg:method} is $(\epsilon,\delta)$-edge DP if the perturbation mechanism in \method{} is $(\epsilon,\delta)$-edge DP.
\end{corollary}
\begin{proof}
    Given that the \method{} ensure $(\epsilon,\delta)$-edge DP, DP guarantee of the trained GCN model with the perturbed adjacency matrix directly follows the post-processing rule \cite{DPFoundation}. 

    As we assume node features and the adjacency matrix have independent distributions, node features will not contain the adjacency information. According to the post-processing rule, any function with the adjacency matrix as input will not increase privacy leakage. Therefore, by regarding the trained GCN model as a function, output from the model still ensure $(\epsilon,\delta)$-edge DP.
\end{proof}

\begin{remark}
\textbf{Inference Privacy. } In transductive settings, the model is fed with a sub-graph from the same graph as in the training stage. Therefore, Algorithm \ref{alg:method} is $(\epsilon,\delta)$-edge DP during inference. In inductive settings, as the model is fed with a graph different from the one in the training stage, \method{} applies the same perturbation as in Algorithm \ref{alg:method}. Therefore, \method{} still ensures $(\epsilon,\delta)$-edge DP. 
\end{remark}

\subsection{Privacy Protection in Practice}\label{sec:privacy:real} 

\subsubsection{Privacy Protection via Low-Rank Decomposition.} 
Besides the DP guarantee by perturbing singular values, \method{} provides additional empirical protection via low-rank decomposition. As stated in Sec \ref{sec:method:decomp}, \method{} trains GNNs using a low-rank graph, $A_{\text{lr}}$, which only consists of the first $r$ principal components in the original graph. As a result, $A_{\text{lr}}$ only consists of the primary topology, while discarding the rest of the connections (See more empirical studies in Sec \ref{sec:exp:ablation}).  Owing to such a low-rank decomposition, it is intuitive to observe that \method{} protects connections that are not included in the low-rank graph, $A_{\text{lr}}$. Meanwhile, \method{} leverages the primary topology in $A_{\text{lr}}$ to train GNNs.

\subsubsection{Privacy Protection beyond Assumption}\label{sec:privacy:unreal}

The establishment of Theorem \ref{theorem:dpsvd} relies on the assumption that $A$ and $A'$ share principal bases. In practice, the effectiveness of the DP guarantee is conditioned by how closely the assumption aligns with the characteristics of real datasets.  

To determine if a graph aligns with the assumption, we can get the singular value matrix, $S'$, from $A'$ using the projection in Eq(\ref{eq:S1}), and compare the magnitude of off-diagonal values in $S'$ with its diagonal values. We can further adopt the Monte Carlo method to randomly delete/add one edge in $A$ for better computation efficiency. 
In the worst case among all $A'$, if the diagonal values are larger than the off-diagonals by several orders of magnitude, the graph and its adjacent will share very similar principal bases. \method{} will provide strong privacy protection by using top $r$ principal bases and perturbing the corresponding singular values. 

The assumption may fail in extreme cases (See Appendix \ref{app:assume} for one extreme example of this case). In such a scenario, the DP guarantee may not hold. For real-world datasets used in this paper, such extreme cases, that adding/deleting one edge results in completely different principal bases, were non-existent. 

\begin{remark}
    Prior work \cite{Stewart1990PerturbationTF} provides theoretical bounds on the difference in singular values and principal bases (Theorem 2 and 4) given small perturbation on a matrix. In \method{}, adjacency matrices only differ at one edge. Hence, the actual difference in singular values and principal bases is within the bound established in \cite{Stewart1990PerturbationTF}. We leave theoretical justification on the assumption of invariance of principal bases for future work.
\end{remark}

\section{Empirical Evaluation}

\subsection{Evaluation Setup}
\textbf{Datasets.} We use 6 standard datasets for the node classification task used in prior work \cite{DBLP:conf/iclr/KipfW17, rozemberczki2021multi, 9833806}. In the transductive setting, we use Cora, Citeseer, Pubmed \cite{sen2008collective}, Chameleon, and Facebook \cite{rozemberczki2021multi} datasets. Facebook is a social network where nodes represent Facebook pages (sites) and the links represent mutual likes between sites. Node features are extracted from the page descriptions provided by the site owners. Cora, Citeseer, Pubmed, and Chameleon are citation networks where the nodes are documents and the edges are citations. The features represent the existence/non-existence of certain keywords. In the inductive setting, we use Twitch dataset \cite{rozemberczki2021multi} which is a collection of 6 disjoint social networks. In Twitch dataset, nodes represent Twitch streamers from different countries and edges represent their mutual friendships. The features for each streamer are based on the games played and streaming habits. Across all graphs, the node features have the same dimension and semantic meaning. The task is to classify whether a node (streamer) uses explicit language. We train GNNs on the graph corresponding to Spain and use other graphs for testing as done previously. Table \ref{tab:dataset} summarizes the statistics of the datasets. We follow the same data splits for training for Cora, Citeseer, Chameleon, and Facebook as considered by the previous works \cite{DBLP:conf/iclr/KipfW17, yang2016revisiting}. For Cora, Citeseer, and Facebook, the training set consists of randomly chosen 20 labels per class, the test set consists of 1000 nodes and the validation set consists of 500 labeled examples. Chameleon comes with 10 random splits of the nodes with 48\% of them used for training, 32\% for validation, and 20\% for testing. The entire graph structure is used for training. For Twitch we do not use any validation set, instead, we use the models with the lowest training loss. 
\begin{table}
  \caption{Overview of dataset statistics.}
  \label{tab:dataset}
    \begin{tabular}{ccccc}
        \toprule
        Dataset&Nodes&Edges&Features&Classes\\
        \midrule
        Cora & 2,708 & 5,429 & 1,433 & 7\\
        Citeseer & 3,327 & 4,732 & 3,703 & 6\\
        Chameleon & 2,277  & 36,101 & 2,325 & 5 \\
        PubMed   & 19,717   & 44,338 & 500  & 3 \\
        Facebook & 22,470 & 171,002 & 128 & 4\\
        \bottomrule
`   \end{tabular}
 \hfill
    \begin{tabular}{ccccc}
        \toprule
        Dataset&Nodes&Edges&Features&Classes\\
        \midrule
        Twitch-ES & 4,648 & 59,382 & 3,170 & 2\\
        Twitch-RU & 4,385 & 37,304 & 3,170 & 2\\
        Twitch-DE & 9,498 & 153,138 & 3,170 & 2\\
        Twitch-FR & 6,549 & 112,666 & 3,170 & 2\\
        Twitch-ENGB & 7,126 & 35,324 & 3,170 & 2\\
        Twitch-PTBR & 1,912 & 31,299 & 3,170 & 2\\
      \bottomrule
    \end{tabular}
\end{table}
\input{plot/privacytradeoff}
\input{plot/privacytradeoff_ind}

\textbf{Model Architectures.} We have 4 model types for all our experiments: GCN (without any privacy as a roofline for model performance), DPGCN, MLP, LPGNet, and \method{}. 
For GCN, we adopt the standard architecture as described in Section \ref{prelim:gnn} along with ReLU activation and dropout for regularization in every hidden layer. For MLP, we use a fully-connected neural network with ReLU activation and dropout in every hidden layer. MLP is by design an edge privacy preserving model since MLP does not use adjacency matrix in the training process and instead relies purely on learning from node features. For LPGNet, we use the same model as in LPGNet \cite{9833806} and choose the LPGNet-2 variant which uses 2 hidden layers and shows higher model accuracy; dropout rate and size of the hidden layers are hyperparameters that can be tuned. For \method{}, we adopt the same architecture as GCN with the adjacency matrix perturbed according to procedures as described in Section \ref{sec:method:training}. Specifically, we set rank equal to 20, and further investigate the effect of rank in Section \ref{sec:exp:rank}. 
We also compare \method{} with the most recent differentially private graph synthesis method PrivGraph, and the results are deferred to Appendix \ref{app:privg}.
We find the right hyperparameters by performing a grid search over a set of possible values in the non-DP setting. We choose the following values for hyperparameter tuning for all models. 
\begin{itemize}
    \item Learning rate: [0.001, 0.005, 0.01, 0.05]
    \item Hidden layer size: [16, 64, 256]
    \item \# Hidden layers: [2, 3] for GCN, \method{} and all MLPs.
    \item Dropout: [0.1, 0.3, 0.5, 0.8]
\end{itemize}
Further, we use two kinds of normalization methods of adjacency matrices
for GCN. For Twitch, we use the First Order GCN normalization technique similar to previous works \cite{rong2019dropedge}(See Eq(\ref{eq:Anormalization1})). For all other datasets, we use the Augmented Normalized Adjacency technique (See Eq(\ref{eq:Anormalization2})). 

In the DP setting, we evaluate using the optimal hyper-parameters obtained from the non-DP setting. 

\textbf{Training and Evaluation Details.} We adopt cross-entropy loss and optimize using Adam’s optimizer with weight decay. For all datasets (except Twitch), we train the models, for 500 epochs and pick the checkpoint that gives the best performance on the validation set during training. This helps us measure the privacy leakage for the model with the best utility. 
For Twitch, we follow the same training procedure as in LINKTELLER paper\cite{9833806} and LPGNet paper \cite{10.1145/3548606.3560705} with $200$ epochs. 
We evaluate the node classification performance using the micro-averaged F1 score following previous works \cite{graphsage,graphsaint} for all datasets except Twitch. 
During inference, node features of a set of test nodes and the adjacency matrix of the test graph (which is the same as the training graph in transductive setting, or different from the training graph in inductive setting) are input into the trained model. The model then outputs the prediction probability vector and selects the class with the highest probability as the predicted label for each test node. The accuracy (F1 score) is computed by comparing the predicted label against the ground-truth label of the test nodes.
For the Twitch dataset, as proposed by the LINKTELLER paper, a modified F1 score is used, which computes the F1 score for the rare class due to significant class imbalance in the dataset. As the task on Twitch dataset is binary classification, we identify the class with fewer data samples as the rare class and consider the rare class as the positive class when calculating the F1 score. 
We run the training for 5 seeds for all models and report the averaged results.

\textbf{Attack Overview.} We evaluate \method{}'s resilience against two attacks: LINKTELLER \cite{9833806} and LPA \cite{he2021stealing}. 
LINKTELLER establishes that predictions of two connected nodes are more easily influenced by changes in one node's features, compared to the nodes that are not connected. The attacker has access to a set of node features and their labels which are required during training. The attacker’s capability also includes the query access to a blackbox GNN model and the obtained prediction probability for a set of nodes during inference. The attacker then computes the \emph{influence} by perturbing the features of a target node and observing the changes in the prediction probability vector of all other nodes from the GNN's outputs. 
With the influence on all nodes,  LINKTELLER can reveal the exact edges of the graph used during inference. As a result, GCNs and other GNN architectures that use graph edges as input are vulnerable to LINKTELLER attacks,  that capture the influence propagated through the edges of the graph for such GCNs. \\
Another attack we consider is LPA. LPA is a more powerful attack that makes use of node features and node embedding correlations to conduct the attack \cite{he2021stealing} and infer the target graph structure. 
It is based on the assumption that nodes with more similar features tend to be connected, which is true for homophily graphs. 
LPA has several different types of attacks based on the knowledge accessible to the attackers, such as node attributes, partial graphs, and a shadow dataset. In our settings where node features are public while the graph structure is completely private, we adopt the LPA attack mode that infers node connections solely based on node features.
Specifically, the LPA attack queries the target model with the known node features and computes the distance between posteriors (i.e., model output probability vectors). Detailed distance metrics can be found in Table 13 in \cite{he2021stealing}. A smaller distance between two nodes' posteriors indicates a higher likelihood of having an edge between them. 
We follow the protocols of both attacks and reproduce them in our experiment.

\textbf{Attack Settings.} We follow the same attack procedures set by LPA and LINKTELLER. For transductive settings, we evaluate the attacks by how well they can classify existing edges and non-edges from a randomly sampled sub-graph. Specifically, we construct a balanced set by sampling an equal number of edges and non-edges ($500$ in our experiments). For inductive settings, we sample a set of 500 inference nodes that are not seen previously during training, and create a subgraph. We then evaluate the attacks on all the edges and non-edges in the subgraph. Note that the number of edges and non-edges can be different in the inductive setting. 
To understand their attack on node degree distribution, as done in LINKTELLER, we also break down the attack performance into low-degree and high-degree nodes and report the results in Section \ref{sec:exp:attack}. \\ 
We use Area Under Receiver Operating Curve (AUC) to measure the attack performance, which is used in LPA and LINKTELLER works. For graph-structured data, the current methods measure AUC by how well an attack can separate a randomly sampled edge from a non-edge. Higher AUC indicates that the attack has a higher success rate of identifying the edges used by the model, or equivalently, the model has lower resilience to the attack. We report the average performance of each attack across 5 runs with different seeds.
\footnotetext{On Twitch-DE, utility of MLP and GCN are the same and close to that of LPGNet.}
\subsection{Privacy-Utility Tradeoff of \method{}}\label{sec:exp:utility}
We first show the privacy-utility tradeoff of \method{} in transductive and inductive settings. 
We train GCN models with different privacy budgets $\epsilon$, ranging from $0.1$ to $8$. Small $\epsilon$ indicates strong privacy constraints. 
Besides the baseline methods (i.e., DPGCN, LPGNet), we also train GCNs with the original adjacency matrix, and MLPs without using the adjacency information at all.

\subsubsection{Transductive Setting}
Figure \ref{fig:f1} shows model utility (F1 score) versus privacy budget ($\epsilon$) in transductive setting.
We observe that 

\begin{enumerate}

\item Compared to DPGCN and LPGNet, \method{} achieves much better model utility under strong privacy constraints. For instance, given $\epsilon<1$, both DPGCN and LPGNet suffer drastic performance drops on all datasets (except Facebook). However, \method{}'s performance is not significantly affected. 

\item Under strong privacy constraints with $\epsilon<1$, \method{} still achieves better model utility compared to MLP that does not use adjacency information during training. The observations indicate that \method{} is very effective in preserving graph information under high privacy constraints, leading to overall performance improvement. 

\item When the epsilon value is large ($\epsilon\geq8$), DPGCN achieves better model utility than \method{}. However, prior works like LPGNet \cite{10.1145/3548606.3560705} (Sec 3) demonstrate that the graph in DPGCN is barely perturbed with large epsilons and essentially the same as the original graph. Thus large epsilons offer little privacy protection in the case of DPGCN. In contrast, \method{} protects privacy by low-rank decomposition and DP noise on singular values. With large epsilons, DP noise fades. However, low-rank decomposition perturbs the graph, and the final model performance is capped at accuracy using a low-rank graph without DP.

\end{enumerate}

Note that for the Facebook dataset, since it is a high homophily graph, nodes with similar features tend to be connected. Therefore, all privacy-preserving training methods give marginal performance improvement by using the edge information from the adjacency matrix. 
Similar observations also hold on PubMed. Even with the full adjacency matrix, the model utility is marginally improved compared to MLP without using the adjacency matrix.
Methods like LPGNet, though, also give slightly better model utility, and significantly compromise protections on edges. 
On the other hand, \method{} can bring much-improved privacy protection, with only slightly affecting the model utility, which is more useful in practice.
\subsubsection{Inductive Setting}
Figure \ref{fig:f1_ind} shows model utility versus privacy budget in inductive setting. \method{} gives a consistent performance under different privacy budgets, which is similar to its performance in transductive setting. 
However, for LPGNet, due to the highly compressed graph representation, its performance is capped even under moderate privacy constraints, different from that in transductive setting. 
Surprisingly, the accuracy of DPGCN is stable when privacy budget varies, and even, for Twitch-FR,  the accuracy increases when privacy budget becomes smaller, which is in sharp contrast to its performance in transductive setting. This may happen because adding noise to edges may generalize models better for DPGCN. We believe Twitch-FR benefitted from this noise. 
However, Twitch-DE is an exception where DPGCN's performance quickly collapses with decreasing $\epsilon$. 
In addition, note that the utility dip at $\epsilon = 6$ is observed on Twitch-ENGB/FR/PTBR in inductive setting except for Twitch-DE dataset. Similar observations are also present in DPGCN \cite{9833806} (Fig 3(a)).
\method{} targets a setting where edge information is private during training and testing. We also measure accuracy when models are trained using \method{} but test graphs are unperturbed, and provide the detailed results in Appendix \ref{app:unperturbed}.

\input{plot/privacyattack}

\subsection{Resilience Against Privacy Attacks}\label{sec:exp:attack}
We further evaluate \method{} against two link stealing attacks: LPA and LINKTELLER attacks in both transductive and inductive settings. To investigate the model utility versus the attack performance, we adjust the privacy parameter $\epsilon$ and respectively evaluate the model's utility and the attacker's performance.

\subsubsection{Transductive Setting}\label{sec:exp:lpa}

For LPA attack, the top row in Figure \ref{fig:attack} shows the model utility versus the LPA attack performance in transductive setting. We choose LPA because it infers edges based on the correlation of node features. Note that different edge perturbations can affect LPA result, because LPA uses similarity of posterior rather than raw node features, to analyze potential edges. The posterior is affected by node features, and aggregation operation in GCN model given edge information. Hence, LPA’s results depend on the model and perturbation methods on edges. We also perform edge attack using raw node feature similarity, and provide the detailed results in Appendix \ref{app:rawnode}. 
As shown in the top row in Figure \ref{fig:attack}, the LPA attack gives an AUC higher than 0.5 (i.e. random guess) for MLP which does not use adjacency matrix at all. 
It indicates that node features and edges share certain common graph information. Therefore, certain connectivity can be inferred via node features. Nevertheless, \method{}, as well as the baseline methods, aim to protect additional information in the adjacency matrix.  

Importantly, \method{} shows stronger resilience against LPA attacks compared to DPGCN and LPGNet. Specifically, we observe that \method{} gives a lower AUC under the same F1 score compared to other baselines. On the other hand, if restricting the attack AUC, \method{} also achieves a higher F1 score. For instance, on the Cora dataset, with an attack AUC of $0.8$, \method{} improves the model utility by $9\%$ compared to DPGCN, and $4\%$ compared to LPGNet.

Note that for the Facebook dataset, due to its high homophily property, the LPA attack recovers most edges by analyzing node features. This indicates that the adjacency matrix of the Facebook graph only encodes common information about the graph, as can be inferred from node features. That is, the adjacency matrix provides little additional \emph{private} information about the graph. 
For the PubMed dataset, as even the full adjacency matrix provides limited performance gains, we mainly focus on edge protection while slightly trading off model utility.

For LINKTELLER attack, the bottom row in Figure \ref{fig:attack} shows the model utility versus the LINKTELLER attack performance. Compared to DPGCN, \method{} improves resilience against the LINKTELLER attack while maintaining the model utility. 
On the other hand, LPGNet seems immune to LINKTELLER attack due to no raw graph information used in the model training. However, it does not indicate that LPGNet is effective in protecting the graph information. On the contrary, it is due to the fact that the LINKTELLER attack is limited as opposed to other attacks, such as the LPA attack, that fully explore node features to improve the attack performance.
As shown in the top row, with additional knowledge from node features, LPGNet quickly loses its capability of protecting the graph.

\subsubsection{Attack Performance Breakdown in terms of Low-Degree and High-Degree Nodes.}\label{sec:exp:ablation}
\pgfplotstableread[col sep=comma]{low.csv}{\loadedtableA}
\pgfplotstableread[col sep=comma]{high.csv}{\loadedtableB}

\begin{figure}[!htb]
    \centering
    \begin{subfigure}[b]{0.23\textwidth}
    \centering
    \begin{tikzpicture}
        \begin{axis} [
            ybar=1pt, bar width=2pt, ticklabel style={font=\scriptsize},
            width=1.2\linewidth, height=.8\linewidth,
            xmin=-0.1, xmax=1.05,
            xtick={0.1, 0.2, 0.4, 0.6,0.8, 0.9}, 
            xticklabels={0.1, 0.2, 0.4, 0.6,0.8, 0.9},
            xlabel={\scriptsize percent of degree changes},
            ylabel={\scriptsize \#nodes}, ymin=0.2, ymax=1400,
            ytick={0, 100, 500, 1000}, yticklabels={0, 100, 500, 1000},
            ymajorgrids, major grid style={line width=.2pt, draw=black!20, dashed},
            legend columns=1,
            legend style={at={(1.1,1.0)}, anchor=north },
        ]
            \addplot[fill=blue!30, draw=blue!60] table [x=bin, y=cnt, col sep=comma] {\loadedtableA};
            
        \end{axis}
    \end{tikzpicture}   
    \caption{\footnotesize low degree nodes}
    \label{fig:deg_lo}
    \end{subfigure}
    \hfill
    \begin{subfigure}[b]{0.23\textwidth}
    \centering
    \begin{tikzpicture}
        \begin{axis} [
            ybar=1pt, bar width=2pt, ticklabel style={font=\scriptsize},
            width=1.2\linewidth, height=.8\linewidth,
            xmin=-0.1, xmax=1.05,
            xtick={0.1, 0.2, 0.4, 0.6,0.8, 0.9}, 
            xticklabels={0.1, 0.2, 0.4, 0.6,0.8, 0.9},
            xlabel={\scriptsize percent of degree changes},
           ymin=0.2, ymax=650,
            ytick={0, 100, 300, 500}, yticklabels={0, 100, 300, 500},
            ymajorgrids, major grid style={line width=.2pt, draw=black!20, dashed},
            legend columns=1,
            legend style={at={(1.1,1.0)}, anchor=north },
        ]
            \addplot[fill=blue!30, draw=blue!60] table [x=bin, y=cnt, col sep=comma] {\loadedtableB};
            
        \end{axis}
    \end{tikzpicture}
    \caption{\footnotesize high degree nodes}
    \label{fig:deg_hi}
    \end{subfigure}
    \vspace{-2mm}
    \caption{Node degree change on Cora dataset (1621 low degree nodes, 1087 high degree nodes). Close to 80\% of low-degree nodes have their node degree modified by 95\% or more, while only around 54\% of high-degree nodes have their node degree changed by at least 95\%.}
    \label{fig:deg}
\end{figure}
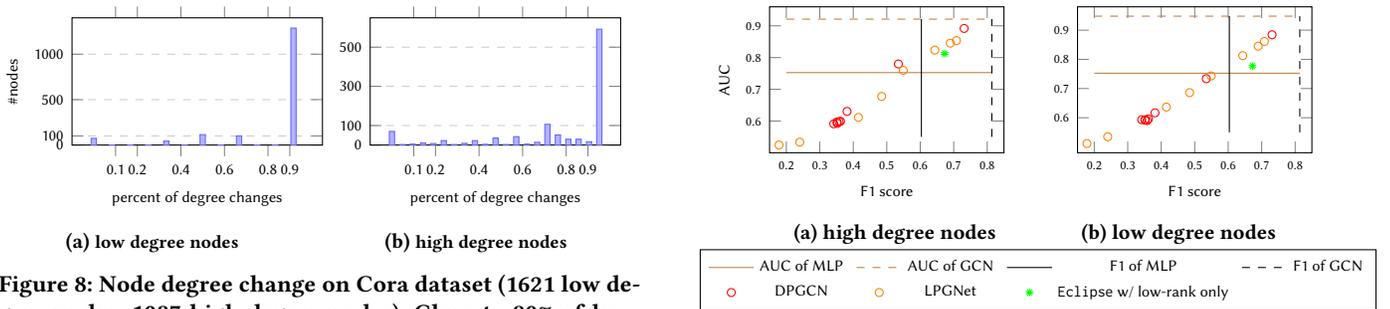
We analyze the effect of low-rank decomposition on nodes with different degrees. 
Following prior works, low (or high) degree nodes refer to nodes with node degree no larger than (or no smaller than) a threshold value $d_{\text{low}}$ (or $d_{\text{high}}$). 
The values $d_{\text{low}}$ and $d_{\text{high}}$ are chosen empirically based on the graph. 
We take Cora dataset as an example and follow the threshold values set for Cora in prior works, where $d_{\text{low}}=3$ and $d_{\text{high}}=4$. Based on these threshold values, $1621$ out of $2708$ nodes in Cora are low-degree nodes, and the remaining $1087$ nodes are high-degree nodes. 
Figure \ref{fig:deg} shows the histogram of node degree change for nodes with different node degrees on the Cora dataset. The percentage change in node degree is plotted on the x-axis, and the number of nodes in each bin is shown on the y-axis. 
With the low-rank decomposition, $80\%$ of low-degree nodes have their node degree reduced by 95\% or more, while only around $54\%$ of high-degree nodes are highly affected. 
Therefore, the low-rank reconstruction already removes substantial number of edges before the DP perturbation. Importantly, it causes more impact on low-degree nodes. 
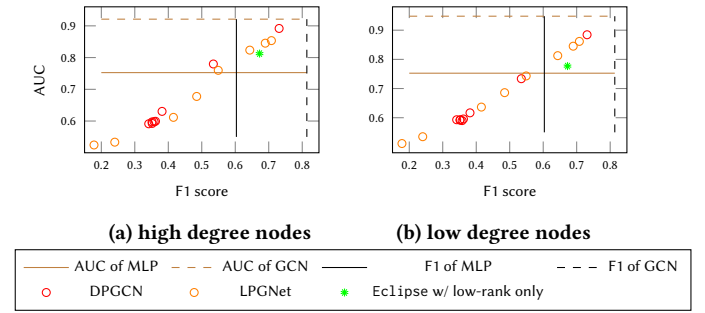
\begin{figure}[!htb]
    \centering
    \begin{subfigure}[!htb]{.22\textwidth}
    \centering
    \captionsetup{justification=centering, margin={10mm, 0mm}}
    \begin{tikzpicture}
        \begin{axis}[
            width=1.2\linewidth, height=.9\linewidth,
            xmin=0.15, xmax=0.85, xlabel={\scriptsize F1 score}, xtick={0.2, 0.3, 0.4, 0.5, 0.6, 0.7,0.8}, xticklabels={0.2, 0.3, 0.4, 0.5, 0.6, 0.7,0.8},
            ymin=0.5, ymax=0.96, ylabel={\scriptsize AUC}, ytick={0.6, 0.7, 0.8, 0.9}, yticklabels={0.6, 0.7, 0.8, 0.9}, 
            major grid style={line width=.2pt, draw=black!30},
            ticklabel style={font=\tiny},
        ]

            \addplot[ brown ] coordinates {
                (0.2, 0.7526) (0.8138, 0.7526)
            };
            \addplot[ brown, dashed ] coordinates {
                (0.2, 0.9211) (0.8138, 0.9211)
            };
            \addplot[ black ] coordinates {
                (0.6034, 0.55) (0.6034, 0.9211)
            };
            \addplot[ black, dashed ] coordinates {
                (0.8138, 0.55) (0.8138, 0.9211)
            };
    
            \addplot+[ 
                only marks, mark=o, red, mark options={solid, scale=0.75}
            ] 
            table [x=x, y=y, col sep=comma]{
                x,        y     
                0.3416	,	0.5913	,
                0.352	,	0.5925	,
            
                0.3516	,	0.5972	,
              
                0.3578	,	0.5978	,
                0.3618	,	0.5992	,
               
                0.3814	,	0.6304	,
             
                0.5346	,	0.7798	,
               
                0.731	,	0.8918	,
              
            };  
              
              
          
          
           
         
            \addplot+[ 
                only marks, mark=o, orange, mark options={solid, scale=0.75}
            ] 
            table [x=x, y=y, col sep=comma]{
                x,        y     
                0.1782	,	0.5245	,
                0.2402	,	0.5334	,
              
                0.4154	,	0.6116	,
              
                0.4848	,	0.6775	,
                0.5492	,	0.7598	,
               
                0.6434	,	0.8235	,
             
                0.6896	,	0.8454	,
             
                0.708	,	0.8535	,
               
            }; 
            \addplot+[ 
                only marks, mark=10-pointed star, green, mark options={solid, scale=0.75}
            ] 
            table [x=x, y=y, col sep=comma]{
                x,        y     
                0.6726	,	0.8126	,
            };
        \end{axis}
    \end{tikzpicture}
    \caption{high degree nodes}
    \label{attack:cora_hinode}
    \end{subfigure} \hspace{2mm}
    \begin{subfigure}[!htb]{.22\textwidth}
    \centering
    \captionsetup{justification=centering}
    \begin{tikzpicture}
        \begin{axis}[
            width=1.2\linewidth, height=.9\linewidth,
            xmin=0.15, xmax=0.85, xlabel={\scriptsize F1 score}, xtick={0.2, 0.3,0.4, 0.5, 0.6, 0.7,0.8}, xticklabels={0.2,0.3,0.4, 0.5, 0.6, 0.7,0.8},
            ymin=0.48, ymax=0.98, ylabel={}, ytick={0.6, 0.7, 0.8, 0.9}, yticklabels={0.6, 0.7, 0.8, 0.9}, 
            major grid style={line width=.2pt, draw=black!30},
            ticklabel style={font=\tiny},
            legend columns=4,
            legend style={/tikz/every even column/.append style={column sep=1mm}, font=\scriptsize},
            legend entries={ AUC of MLP, AUC of GCN, F1 of MLP, F1 of GCN, DPGCN, LPGNet, \method{} w/ low-rank only},
            legend to name=named_node,
        ]

            \addplot[ brown ] coordinates {
                (0.2, 0.7526) (0.8138, 0.7526)
            };
            \addplot[ brown, dashed ] coordinates {
                (0.2, 0.9479) (0.8138, 0.9479)
            };
            \addplot[ black ] coordinates {
                (0.6034, 0.55) (0.6034, 0.9479)
            };
            \addplot[ black, dashed ] coordinates {
                (0.8138, 0.55) (0.8138, 0.9479)
            };
    
            \addplot+[ 
                only marks, mark=o, red, mark options={solid, scale=0.75}
            ] 
            table [x=x, y=y, col sep=comma]{
                x,        y     
                0.3416	,	0.5931	,
                0.352	,	0.5932	,
             
                0.3516	,	0.5917	,
               
                0.3578	,	0.5902	,
                0.3618	,	0.5957	,
             
                0.3814	,	0.6168	,
              
                0.5346	,	0.7337	,
              
                0.731	,	0.8843	,
             
            };  
               
            
        
            
             
                
            \addplot+[ 
                only marks, mark=o, orange, mark options={solid, scale=0.75}
            ] 
            table [x=x, y=y, col sep=comma]{
                x,        y     
                0.1782	,	0.5116	,
                0.2402	,	0.535	,
         
                0.4154	,	0.6365	,
             
                0.4848	,	0.6859	,
                0.5492	,	0.7432	,
           
                0.6434	,	0.8127	,
         
                0.6896	,	0.8451	,
            
                0.708	,	0.8615	,
             
            }; 
            \addplot+[ 
                only marks, mark=10-pointed star, green, mark options={solid, scale=0.75}
            ] 
            table [x=x, y=y, col sep=comma]{
                x,        y     
                0.6726	,	0.7769	,
            };
        \end{axis}
    \end{tikzpicture}
    \caption{low degree nodes}
    \label{attack:cora_lonode}
    \end{subfigure}

    \ref{named_node}
    \vspace{-2mm}
    \caption{LPA Attack AUC vs. Utility F1 score. in transductive setting breakdown for low and high degree nodes in the Cora dataset.}
    \label{fig:attack_node}
\end{figure}
\input{plot/privacyattack_ind}
\input{plot/privacyattack_Rank}
Figure \ref{fig:attack_node} shows a detailed LPA attack performance on nodes of different node degrees. We only show LPA attack performance because it is a stronger attack on edges than LINKTELLER attack as discussed in Sec. \ref{sec:exp:lpa}. 
First, \method{} with only a low-rank graph (no DP perturbation) already provides edge protection. Importantly, for low- and high-degree nodes, \method{} without DP achieves better model utility than other baselines given the same attack AUC. The reason is that the low-rank decomposition in \method{} preserves the primary topology for better model utility and removes secondary edges in graphs for edge protection.
We further observe that LPA attack performs better on high-degree nodes as expected, as a higher percentage of edges from high-degree nodes are preserved in the low-rank adjacency matrix. 
However, for low-degree nodes, the attack performance is close to the performance of MLP, which does not use the adjacency matrix during training.  
Therefore, the results show \method{} provides stronger protection on low-degree nodes. For real-world graph data, a high-degree node denotes a centroid in a graph, like a well-known person in social networks. Common DP mechanisms are usually effective in protecting the connectivity of high-degree nodes. However, low-degree nodes are hard to protect \cite{287254}. \method{} fills this gap by filtering out connections of low-degree nodes with low-rank reconstruction.
\subsubsection{Inductive Setting}
Figure \ref{fig:attack_ind} shows the attack performance on low-degree nodes in inductive setting.
We observe that for most datasets, \method{} achieves much better model utility while still maintaining similar resilience against LPA and LINTELLER attacks. 
Specifically, compared to DPGCN, \method{} archives comparable model utility but with much lower attack AUC, indicating that \method{} achieves a much better tradeoff between model performance and edge privacy leakage. 
Unlike LPGNet, which sacrifices significant model performance with low AUC, model performance using \method{} reaches close to that of GCN with the original adjacency matrix.
However, Twitch-RU is an exception where model utility of \method{} is lower than both DPGCN and LPGNet given similar attack AUC. This may happen because Twitch-RU graph has lower node degree(1-3 edges on average) than other graphs. The low rank clipped many edges causing the graph to lose some topology and thus lower the model utility.

\subsection{Effect of Rank}\label{sec:exp:rank}
In this experiment, we investigate how changing the rank affects the accuracy/privacy performance of our proposed method \method{}. We vary rank from 10 to 200 and report attack AUC vs. F1 score under 5 different privacy budgets $\epsilon \in \{0.1,0.5,1,5,10\}$. 
We run the target model and the attack model for 5 seeds. The results are shown in Figure \ref{fig:attack_rank}. We observe that as rank increases,  model utility improves. On the other hand, the AUC of LPA and LINKTELLER attack also increases across all three datasets in transductive setting. 
The reason is that, as we increment rank, more edges are included in the low-rank graph, leading to potential leakage after DP perturbation. 
Specifically, during training, more true edges will be involved, resulting in an increasing number of edges in the original graph being reconstructed. As a result, model utility is improved. 
However, more true edges, that are included in the perturbed adjacency matrix, further reveal the node influence and node feature correlations during training and inference, which compromises the trained model's resilience against attacks. 
We also observed that while the attack AUC increases with the rank of the perturbed graph, the attack accuracy never exceeds that of a non-private GCN model, meaning that the low-rank reconstruction still removes some of the information in the original graph effectively.

Based on the observation above, we suggest a practical approach for selecting the appropriate rank to train models using \method{} with reasonable attack AUC. For instance, we set an upper threshold (e.g., 0.65 ) for the attack AUC (e.g., LINKTELLER attack), select the largest possible rank (e.g., 20 for Cora dataset), and train the target model that satisfies the privacy constraint measured by the attack AUC. Then, we evaluate the model utility under the selected rank.

\section{Conclusion}

In this paper, we presented \method{}, a new privacy-preserving GNN training algorithm that ensures edge-level differential privacy for sensitive graph edges. \method{} achieves better privacy-utility tradeoff compared to current state-of-the-art privacy-preserving GNN training methods. 
\method{} trains GNNs with a low-rank format of the graph via singular value decomposition.
The low-rank graph removes many edges from the original graph, while still maintaining enough of the graph structure for better model utility. 
We further use Gaussian mechanism on the low-rank singular values of the graph's adjacency matrix to protect edges in the low-rank graph. 
Extensive experiments on real-world graph datasets show that our method achieves good model utility while providing strong privacy protection on edges and outperforms existing methods. 
We also explored and evaluated the impact of node degree distribution and rank on the attack resilience of our method. \\
In future work, we will explore providing theoretical justification on the assumption of invariance of principal bases. We will also consider a more adaptive rank selection scheme to further improve privacy-utility tradeoffs when training GNNs. While Eclipse targets scenarios where edge information is private, commonly seen in real-world applications \cite{9833806,10.1145/3548606.3560705}, we believe that privacy protection using low-rank decomposition can be extended to node attributes, by exploiting potential low-rank properties in node attributes. 


\begin{acks}
This material is based upon work supported by Defense Advanced Research Projects Agency (DARPA) under Contract Nos.\\HR001120C0088, NSF award number 2224319, REAL@USC-Meta center, and VMware gift. The views, opinions, and/or findings expressed are those of the author(s) and should not be interpreted as representing the official views or policies of the Department of Defense or the U.S. Government.
\end{acks}

\bibliographystyle{ACM-Reference-Format}
\bibliography{PETS2024_main}

\appendix
\section{Proof of Theorem \ref{theorem:dpsvd}}\label{appx:thm1}
\setcounter{theorem}{0}
\begin{theorem}
Given $A$ and $A'$ share the same principal bases, with Gaussian noise,
\begin{center}
    $ \texttt{Normal} \left (0, \Delta\sqrt{2\ln({1.25/\delta})} / \epsilon_{\text{lr}} \right )$,
\end{center} 
added to singular values;
and Laplacian noise, 
\begin{center}
    $\texttt{Laplacian} \left (0,1 / \epsilon_{\text{e}} \right)$,
\end{center}
added to the edge count, the perturbed low-rank $\Aplr$ using \method{} satisfies $(\epsilon,\delta)$-edge DP with $\epsilon = \epsilon_{\text{lr}} + \epsilon_{\text{e}}$.
\end{theorem}
\begin{proof}
The proof follows the standard Gaussian mechanism and the post-processing and composition rule of the difference privacy \cite{DPFoundation}. 
\method{} consists of two randomized mechanisms: A Gaussian mechanism on singular values and a Laplacian mechanism on the edge count. 
First, we show that perturbed singular values $\tilde{s}$ is $(\epsilon_{lr},\delta)$-edge differentially private. \\
Given sensitivity $\Delta$, with Gaussian noise $\texttt{Normal}(0, \frac{\Delta\sqrt{2\ln({1.25/\delta})}}{\epsilon_{lr}})$ adding to singular values, $\tilde{s}$ (line 9 in Algorithm \ref{alg:method}) is $(\epsilon_{lr},\delta)$-edge differentially private based on the standard Gaussian mechanism. \\
When obtaining a low-rank $A_{\text{lr}}$ (line 11 in Algorithm \ref{alg:method}), as it is the post-processing given $\tilde{s}$ and shared principal bases, it does not change the differential privacy budget. Hence, the low-rank reconstruction follows the same privacy budget as $\tilde{s}$. \\
Then, for the Laplacian mechanism on the edge count (line 5 in Algorithm \ref{alg:method}), $\tilde{E}$ satisfied $(\epsilon_e,0)$-edge differentially private. \\ 
Given a low-rank $A_{\text{lr}}$ and the edge count $\tilde{E}$, we apply binary quantization (line 16 in Algorithm \ref{alg:method}) to obtain the final low-rank adjacency matrix $\Aplr$. 
By the composition theorem and the post-processing rule of differential privacy, we obtain the final privacy guarantee 
\begin{center}
    $\epsilon = \epsilon_{lr}+\epsilon_e$, $\quad \delta = \delta + 0$.
\end{center}
Hence, the perturbed low-rank $\Aplr$ using \method{} guarantees $(\epsilon,\delta)$-edge differential privacy. 
\end{proof}

\section{Cosine Similarity of Principal Bases for Other Datasets}\label{app:cosine}
\begin{figure}[!htb]
    \centering
    \begin{tikzpicture}
        \begin{axis} [
            ybar=1pt, bar width=2pt, ticklabel style={font=\scriptsize},
            width=.9\linewidth, height=.4\linewidth,
            xmin=0, xmax=21,
            xtick={1, 5, 10, 15, 20}, 
            xticklabels={1, 5, 10, 15, 20},
            xlabel={\scriptsize Index},
            ylabel={\scriptsize Similarity}, ymin=0.2,
            ytick={0, 0.5, 1}, yticklabels={0, 0.5, 1},
            ymajorgrids, major grid style={line width=.2pt, draw=black!20, dashed},
            legend columns=1,
            legend style={at={(1.1,1.0)}, anchor=north },
            legend entries={\tiny $U$, \tiny $V$}
        ]
            \addplot[fill=blue!30, draw=blue!60] coordinates {
                (1, 0.998) (2, 0.997) (3, 0.997) (4, 0.997) (5, 0.997)
                (6, 0.997) (7, 0.997) (8, 0.997) (9, 0.997) (10, 0.998)
                (11, 0.998) (12, 0.998) (13, 0.998) (14, 0.998) (15, 0.998)
                (16, 0.998) (17, 0.998) (18, 0.998) (19, 0.982) (20, 0.982)
            };
				
            \addplot[fill=magenta!30, draw=magenta!60] coordinates {
                (1, 1.004) (2, 1.005) (3, 1.005) (4, 1.005) (5, 1.005)
                (6, 1.005) (7, 1.005) (8, 1.005) (9, 1.006) (10, 1.005)
                (11, 1.006) (12, 1.006) (13, 1.005) (14, 1.006) (15, 1.006)
                (16, 1.005) (17, 1.006) (18, 1.005) (19, 0.991) (20, 0.991)
            };
            
        \end{axis}
    \end{tikzpicture}   
    \vspace{-3mm}
    \caption{Cosine similarity of principal basis vectors in the Facebook dataset.}
    \label{fig:cos_fb}
\end{figure}
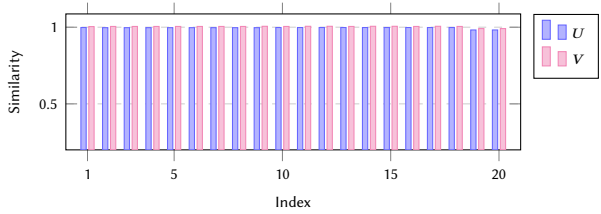

Figure \ref{fig:cos_fb} show the cosine similarity of principal bases between adjacent graphs on the larger Facebook dataset. Similar as Figure \ref{fig:cos}, principal bases between adjacent graphs are almost the same.

\section{Privacy-Utility Tradeoff compared with PrivGraph}\label{app:privg}

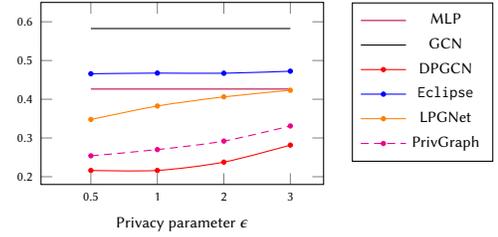
\begin{figure}[!htb]
    \centering
    \begin{subfigure}[!htb]{.5\textwidth}
    \centering
    \begin{tikzpicture}
        \begin{axis}[
            width=.6\linewidth, height=.45\linewidth,
            xmin=0.25, xmax=4.5, xlabel={\scriptsize Privacy parameter $\epsilon$}, xtick={1, 2,3,4}, xticklabels={0.5,1, 2, 3},
            ymin=0.18, ymax=0.65, ytick={0.2, 0.3,0.4,0.5,0.6}, yticklabels={0.2, 0.3,0.4,0.5,0.6}, 
            major grid style={line width=.2pt, draw=black!30},
            ticklabel style={font=\tiny},
            legend columns=1,
            legend style={at={(1.1,1)}, anchor=north west},
            legend entries={\scriptsize MLP , \scriptsize GCN, \scriptsize DPGCN ,\scriptsize \method{} ,\scriptsize LPGNet, \scriptsize PrivGraph},
        ]

            \addplot[ purple ] coordinates {
                (1, 0.4263) (4, 0.4263)
            };
            \addplot[ black ] coordinates {
                (1, 0.5825) (4, 0.5825)
            };
    
            \addplot+[ 
                red, smooth, mark=*, mark options={solid, scale=0.4},
                error bars/.cd, y fixed, y dir=both, y explicit
            ] 
            table [x=x, y=y, col sep=comma]{
                x,        y
                1	,	0.2162	,
                2	,	0.2162	,
                3	,	0.2377	,
                4   ,   0.2816  ,
            };   
            \addplot+[ 
                blue, smooth, mark=*, mark options={solid, scale=0.4},
                error bars/.cd, y fixed, y dir=both, y explicit
            ] 
            table [x=x, y=y, col sep=comma]{
                x,        y
                1	,	0.4658	,
                2	,	0.4675	,
                3	,	0.4671	,
                4   ,   0.4724  ,
            };  
            \addplot+[ 
                orange, smooth, mark=*, mark options={solid, scale=0.4},
                error bars/.cd, y fixed, y dir=both, y explicit
            ] 
            table [x=x, y=y, col sep=comma]{
                x,        y
                1	,	0.3478	,
                2	,	0.3825	,
                3	,	0.4061	,
                4   ,   0.4232  ,
            }; 
            \addplot+[ 
                magenta, smooth, mark=*, mark options={solid, scale=0.4},
                error bars/.cd, y fixed, y dir=both, y explicit
            ] 
            table [x=x, y=y, col sep=comma]{
                x,        y
                1	,	0.254	,
                2	,	0.27	,
                3	,	0.292	,
                4   ,   0.331  ,
            }; 
        \end{axis}
    \end{tikzpicture}
    \end{subfigure}  
    \vspace{-2mm}
    \caption{Model utility on Chameleon dataset. \method{} achieves higher accuracy compared to DPGCN, LPGNet, and PrivGraph, while PrivGraph performs better than DPGCN.}
    \label{fig:f1_privgraph}
\end{figure}

PrivGraph is a graph synthesis algorithm that exploits community information to publish a synthetic graph under DP guarantee. PrivGraph aims to reduce the perturbation noise to the adjacency matrix while limiting the information loss during encoding the graph data, which is similar to the goal of \method{}. To compare \method{}’s perturbation method to PrivGraph, we follow PrivGraph’s procedure to generate a synthetic graph and use it to train a GCN model on Chameleon (the common dataset used in \method{} and PrivGraph). We try 4 different privacy budgets $\epsilon \in \{0.5,1,2,3\}$ following PrivGraph's experimental setup, and report the GCN model’s test accuracy in Figure \ref{fig:f1_privgraph}. We found that PrivGraph outperforms DPGCN which perturbs the adjacency matrix directly. However, given the same privacy budget, PrivGraph achieves lower test accuracy than \method{}. 
One reason is that PrivGraph relies on \emph{community information} when reconstructing edges. However, Chameleon is considered a heterophilous dataset where the features and edges do not correlate well with the ground-truth labels (clusters/communities). The community detection algorithm used in PrivGraph cannot effectively capture connection information for such heterophilous graph. 
On the other hand, \method{} uses the low-rank method to preserve the primary topology, which is independent of how closely edges and community information correlate. 


\section{When will the assumption fail?}\label{app:assume}
Consider a fully connected graph $G_F$ consisting of 3 nodes only, with its adjacency matrix denoted as $A_F$. Upon SVD, we obtain the singular vectors and singular values as shown below. To obtain its neighboring graph $G_F'$, we remove the edge between node 1 and node 2, the resulting adjacency matrix $A_F'$ and its singular vectors and singular values can be found as shown below. We can see that $G_F$ and $G_F'$ do not share principal bases. However, such extreme cases, where adding/deleting one edge results in completely different principal bases, are not observed in real-world datasets used in this paper. 
\begin{equation*}
    A_F \xrightarrow[]{SVD} U_F \cdot S_F \cdot V_F
\end{equation*}
\begin{equation*}
\begin{split}
    &\begin{bmatrix}
    0 & 1 & 1\\
    1 & 0 & 1\\
    1 & 1 & 0
    \end{bmatrix} \xrightarrow[]{SVD} \\
    &\begin{bmatrix}
    0.58 & -0.71 & -0.41\\
    0.58 & 0.71 & -0.41\\
    0.58 & 0 & 0.82
    \end{bmatrix} \cdot
    \begin{bmatrix}
    2 & 0 & 0\\
    0 & 1 & 0\\
    0 & 0 & 1
    \end{bmatrix} \cdot
    \begin{bmatrix}
    0.58 & 0.71 & -0.41\\
    0.58 & -0.71 & -0.41\\
    0.58 & 0 & -0.82
    \end{bmatrix}
\end{split}
\end{equation*}

\vspace{3mm}
\begin{equation*}
    A_F' \xrightarrow[]{SVD} U_F' \cdot S_F' \cdot V_F'
\end{equation*}

\begin{equation*}
\begin{split}
    &\begin{bmatrix}
    0 & 0 & 1\\
    0 & 0 & 1\\
    1 & 1 & 0
    \end{bmatrix} \xrightarrow[]{SVD} \\
    &\begin{bmatrix}
    0.71 & 0 & -0.71\\
    0.71 & 0 & 0.71\\
    0& 1 & 0
    \end{bmatrix} \cdot
    \begin{bmatrix}
    1.41 & 0 & 0\\
    0 & 1.41 & 0\\
    0 & 0 & 0
    \end{bmatrix} \cdot
    \begin{bmatrix}
    0 & 0.71 & -0.71\\
    0 & 0.71 & 0.71\\
    1 & 0 & 0
    \end{bmatrix}
\end{split}
\end{equation*}

\section{Unperturbed Test Graph}\label{app:unperturbed}

\method{} targets a setting where edge information is private during training and testing, commonly seen in edge-DP literature \cite{9833806,10.1145/3548606.3560705}. We also measure accuracy when models are trained using \method{} but test graphs are unperturbed and report the test accuracy in Table \ref{tab:unperturbed}. We use Twitch dataset for inductive settings where training and test graphs differ. We observe that \method{} is generally robust to test graph perturbations. Both unperturbed and perturbed graphs provide similar performance.

\section{Attack using Node Feature Similarity}\label{app:rawnode}

We perform edge attack using raw node feature similarity and report the attack AUC as well as the AUC drop compared to LPA in Table \ref{tab:rawnode}. Not surprisingly using only node features, attack AUC is generally lower than that achieved with LPA (brown-dashed lines Figure \ref{fig:attack}), using model output similarity.  
\begin{table}[!htb]
  \caption{Test Accuracy when test graphs are unperturbed and perturbed.}
  \small
  \label{tab:unperturbed}
    \begin{tabular}{cccc}
        \toprule
        Dataset&$\epsilon$&Acc. w/o perturbation  & Acc. w/ perturbation\\
        \midrule
TwitchDE	&	0.1	&	0.567	&	0.55	\\
	&	0.2	&	0.571	&	0.552	\\
	&	0.5	&	0.569	&	0.555	\\
	&	1	&	0.567	&	0.554	\\
	&	2	&	0.567	&	0.556	\\
	&	4	&	0.567	&	0.556	\\
	&	6	&	0.567	&	0.556	\\
	&	8	&	0.568	&	0.555	\\
        \midrule
TwitchFR	&	0.1	&	0.379	&	0.395	\\
	&	0.2	&	0.378	&	0.397	\\
	&	0.5	&	0.375	&	0.398	\\
	&	1	&	0.374	&	0.386	\\
	&	2	&	0.35	&	0.369	\\
	&	4	&	0.354	&	0.372	\\
	&	6	&	0.349	&	0.373	\\
	&	8	&	0.348	&	0.374	\\
        \midrule
TwitchRU	&	0.1	&	0.226	&	0.245	\\
	&	0.2	&	0.224	&	0.24	\\
	&	0.5	&	0.229	&	0.232	\\
	&	1	&	0.223	&	0.238	\\
	&	2	&	0.221	&	0.233	\\
	&	4	&	0.223	&	0.237	\\
	&	6	&	0.224	&	0.237	\\
	&	8	&	0.227	&	0.241	\\
        \midrule
TwitchENGB	&	0.1	&	0.618	&	0.609	\\
	&	0.2	&	0.618	&	0.609	\\
	&	0.5	&	0.619	&	0.61	\\
	&	1	&	0.619	&	0.611	\\
	&	2	&	0.619	&	0.612	\\
	&	4	&	0.619	&	0.612	\\
	&	6	&	0.619	&	0.612	\\
	&	8	&	0.619	&	0.613	\\
        \midrule
TwitchPTBR	&	0.1	&	0.395	&	0.45	\\
	&	0.2	&	0.396	&	0.455	\\
	&	0.5	&	0.381	&	0.448	\\
	&	1	&	0.363	&	0.443	\\
	&	2	&	0.369	&	0.439	\\
	&	4	&	0.372	&	0.436	\\
	&	6	&	0.374	&	0.442	\\
	&	8	&	0.387	&	0.439	\\
        \bottomrule
   \end{tabular}
\end{table} 
\begin{table}[!htb]
  \caption{Attack AUC using node feature similarity.}
  \label{tab:rawnode}
    \begin{tabular}{ccc}
        \toprule
        Dataset&AUC&AUC drop v.s. LPA\\
        \midrule
        Cora & 0.821 & -0.12 \\
        Citeseer & 0.902 & -0.06 \\
        Chameleon & 0.571  & -0.19 \\
        PubMed   & 0.902   & 0.04 \\
        Facebook & 0.636 & -0.24 \\
        \bottomrule
    \end{tabular}
    \vspace{-5mm}
\end{table}

\end{document}